%% file: main_arxiv.tex
\documentclass[11pt]{article}
\usepackage[utf8]{inputenc}
\usepackage[T1]{fontenc}
\usepackage{lmodern}
\input{command}

\input{notation}
\title{Near Minimax Rates for Quantile Distributional Reinforcement Learning}
\author{
Zijie Cheng\thanks{School of Mathematical Sciences, Peking University; email: \texttt{xmwsbsj@stu.pku.edu.cn}.}
\and
Zhihua Zhang\thanks{School of Mathematical Sciences, Peking University; email: \texttt{zhzhang@math.pku.edu.cn}.}
}
\begin{document}
\maketitle
\input{sections/abstract}
\section{Introduction}\label{sec:intro}\label{Section:intro}
\input{sections/introduction}

\section{Preliminaries}\label{sec:prelim}\label{Section:preliminary}
\input{sections/preliminaries}
\section{Main Results}\label{sec:results}\label{Section:analysis}
\input{sections/results}
\section{Proof Outlines}\label{sec:strategy}\label{Section:proof_outline}
\input{sections/proof_outlines}
\section{Conclusions}\label{sec:conclusion}\label{Section:conclusion}
\input{sections/conclusion}

\bibliographystyle{abbrvnat}
\bibliography{ref}
\clearpage
\appendix
\section{Omitted Proofs of Section~\ref{sec:strategy}}\label{app:key}
\input{appendices/key_lemmas}

\section{Technical Lemmas for Appendices~\ref{app:proof-A3-group} and~\ref{app:proof-A4-Winf-group}}\label{app:basic}
\input{appendices/basic}

\section{Technical Lemmas for Appendix~\ref{app:proof-A4-W2-group}}\label{app:geometry}\label{app:inverse}\label{app:empirical}
\input{appendices/geometry}
\section{Technical Lemmas for Appendix~\ref{app:proof-local-minimax-group}}\label{app:local}
\input{appendices/local}
\section{Technical Lemmas for Appendix~\ref{app:proof-hard-instance-group}}\label{app:regularity}
\input{appendices/construction}
\end{document}

%% file: command.tex
\usepackage{geometry}
\usepackage{setspace}
\usepackage{amsmath, amssymb, amsfonts, bm, mathtools, mathrsfs}
\usepackage{amsthm}
\usepackage[dvipsnames]{xcolor}
\definecolor{darkblue}{rgb}{0,0,.5}
\usepackage{graphicx}
\usepackage{subfigure}
\usepackage[numbers]{natbib}
\usepackage[colorlinks=true,allcolors=darkblue]{hyperref}       
\usepackage{url}            
\usepackage{booktabs}       
\usepackage{amsfonts}       
\usepackage{nicefrac}       
\usepackage{microtype}      

\allowdisplaybreaks

\usepackage{float}
\usepackage{multirow}
\usepackage{footnote}
\IfFileExists{dsfont.sty}{\usepackage{dsfont}}{\providecommand{\mathds}{\mathbf}}
\IfFileExists{mathabx.sty}{\usepackage{mathabx}}{}

\IfFileExists{algorithm.sty}{\usepackage{algorithm}}{}
\IfFileExists{algorithmic.sty}{\usepackage{algorithmic}}{}
\usepackage{nicefrac}
\usepackage{tikz}
\usepackage{overpic}

\IfFileExists{dsfont.sty}{\usepackage{dsfont}}{\providecommand{\mathds}{\mathbf}}
\usepackage{hyperref}
\usepackage[capitalize]{cleveref}
\usepackage{crossreftools}

\makeatletter
\newcommand*{\rom}[1]{\expandafter\@slowromancap\romannumeral #1@}
\makeatother

\newcommand{\ind}{\mathds{1}}
\newcommand{\var}{\mathsf{Var}}
\newcommand{\E}{\mathbb{E}}

\newcommand{\brc}[1]{\left\{{#1}\right\}}
\newcommand{\prn}[1]{\left({#1}\right)} 
\newcommand{\brk}[1]{\left[{#1}\right]} 
\newcommand{\norm}[1]{\left\|{#1}\right\|} 
\newcommand{\abs}[1]{\left|{#1}\right|} 
\newcommand{\what}[1]{\widehat{#1}}
\newcommand{\wtilde}[1]{\widetilde{#1}}

\newcommand{\rd}{\mathrm{d}}
\newcommand{\normal}{\mathsf{N}}

\newcommand{\<}{\langle} 
\renewcommand{\>}{\rangle}

\newcommand{\simiid}{\stackrel{\textup{iid}}{\sim}}

\newcommand{\cd}{\stackrel{d}{\rightarrow}}

\def\gA{{\mathcal{A}}}
\def\gB{{\mathcal{B}}}

\def\gD{{\mathcal{D}}}
\def\gE{{\mathcal{E}}}
\def\gF{{\mathcal{F}}}

\def\gH{{\mathcal{H}}}
\def\gI{{\mathcal{I}}}

\def\gM{{\mathcal{M}}}
\def\gN{{\mathcal{N}}}

\def\gP{{\mathcal{P}}}
\def\gQ{{\mathcal{Q}}}
\def\gR{{\mathcal{R}}}
\def\gS{{\mathcal{S}}}
\def\gT{{\mathcal{T}}}
\def\gU{{\mathcal{U}}}

\def\sP{{\mathscr{P}}}

\def\bA{{\bm{A}}}
\def\bR{{\bm{R}}}

\def\bD{{\bm{D}}}

\def\bH{{\bm{H}}}
\def\bK{{\bm{K}}}

\def\bx{{\bm{x}}}
\def\bI{{\bm{I}}}

\def\bG{{\bm{G}}}

\def\bU{{\bm{U}}}

\def\bq{{\bm{q}}}
\def\bPi{{\bm{\Pi}}}
\def\bphi{{\bm{\phi}}}

\def\bSigma{{\bm{\Sigma}}}
\def\btheta{{\bm{\theta}}}

\def\bu{{\bm{u}}}
\def\bv{{\bm{v}}}
\def\bxi{{\bm{\xi}}}
\def\btheta{{\bm{\theta}}}
\def\bV{{\bm{V}}}
\def\bY{{\bm{Y}}}

\def\bZ{{\bm{Z}}}

\def\bD{{\bm{D}}}
\def\bz{{\bm{z}}}

\def\RB{{\mathbb R}}
\def\EB{{\mathbb E}}

\def\PB{{\mathbb P}}

\def\NB{{\mathbb N}}

\def\HB{{\mathbb H}}

\newcommand{\KS}{\textup{KS}}
\newcommand{\TV}{\textup{TV}}

\makeatletter
\long\def\@makecaption#1#2{
  \vskip 0.8ex
  \setbox\@tempboxa\hbox{\small {\bf #1:} #2}
  \parindent 1.5em  
  \dimen0=\hsize
  \advance\dimen0 by -3em
  \ifdim \wd\@tempboxa >\dimen0
  \hbox to \hsize{
    \parindent 0em
    \hfil 
    \parbox{\dimen0}{\def\baselinestretch{0.96}\small
      {\bf #1.} #2
    } 
    \hfil}
  \else \hbox to \hsize{\hfil \box\@tempboxa \hfil}
  \fi
}
\makeatother

\newcommand{\Beta}{\boldsymbol{\eta}}

\newcommand{\cov}{\mathsf{Cov}}

%% file: notation.tex
\newcommand{\bvartheta}{\bm{\vartheta}}
\newcommand{\Wbar}{\overline W}

\newcommand{\btau}{\bm{\tau}}
\newcommand{\bt}{\bm{t}}
\usepackage{enumitem}
\usepackage{needspace}
\newtheorem{claim}{Claim}[section]
\newtheorem{lemma}[claim]{Lemma}
\newtheorem{assumption}{Assumption}

\newtheorem{theorem}{Theorem}[section]

\newtheorem{proposition}{Proposition}[section]

\newtheorem{corollary}{Corollary}[section]

%% file: sections/abstract.tex
\begin{abstract}
In this paper, we study quantile distributional policy evaluation, where the return distribution under a given policy is approximated by the quantile Bellman fixed point \(\bm\eta_m\). We investigate the statistical accuracy of estimating \(\bm\eta_m\) across quantile resolutions under Wasserstein distances. Assuming access to a generative model, we construct a quantile fixed point estimator $\bm\eta_m^{(n)}$ from an empirical Markov decision process of $n$ observations per state-action pair and establish non-asymptotic error bounds under Wasserstein distances. When reward densities are uniformly bounded away from zero, we obtain a root-\(n\) high-probability bound simultaneously over all resolutions \(m\) and all Wasserstein orders \(1\le p\le\infty\). When reward densities vanish at the support endpoints, we show that continuity and endpoint monotonicity suffice to preserve the same resolution-independent rate for \(1\le p\le2\). For \(p>2\), we establish a root-\(n\) bound with coefficient \(m^{1/2-1/p}\), together with a slower, resolution-independent \(n^{-1/p}\) bound for every finite \(p>2\). We further obtain resolution-independent \(W_\infty\) consistency governed by the continuity modulus of reward quantile functions.
To investigate the sharpness of these bounds, we characterize the exact local asymptotic minimax risk at each fixed resolution as the expected Wasserstein-coordinate norm of the efficient Gaussian limit and prove that our estimator attains it. We then examine the resolution dependence of this optimal local risk. A regular one-state instance has resolution-independent local risk constants, whereas a single endpoint-vanishing two-state instance exhibits growth of order \(m^{1/2-1/p}\), up to logarithmic factors along a resolution subsequence, for every \(p>2\).
\end{abstract}

%% file: sections/introduction.tex
Reinforcement learning has made remarkable advances in a variety of applications, ranging from game playing \citep{silver2018general,vinyals2019grandmaster} and robotics \citep{kober2013reinforcement} to large language models \citep{ouyang2022training,openai2023gpt4}. In the classical framework motivated by the reward hypothesis \citep{sutton2004,sutton2018reinforcement}, a policy is evaluated through its expected returns, namely the expected discounted cumulative reward.
However, this criterion alone does not capture the variability and tail risks of returns, which are crucial in many real-world applications. 
For example, evaluating dynamic treatment regimes requires attention to adverse patient outcomes beyond average treatment effects \citep{lavori2004dynamic}. In finance, investors face a trade-off between expected returns and risk, as higher expected returns are often accompanied by greater uncertainty \citep{ghysels2005there}.

Distributional reinforcement learning \citep{morimura2010nonparametric,bellemare2017distributional} extends conventional reinforcement learning by modeling the entire return distribution instead of its expectation. By accounting for randomness in both the environment and the policy, this approach offers a more complete description of return variability and uncertainty. 
However, since return distributions are generally infinite-dimensional, practical algorithms rely on finite-dimensional approximations, with categorical and quantile representations being two major approaches. Introduced by \citet{dabney2018distributional}, quantile-based methods approximate return distributions using finitely many quantiles, providing a flexible and computationally tractable representation that retains important distributional information. 
As a result, quantile-based approaches have become widely used in modern distributional reinforcement learning, including the quantile regression deep Q-network (QR-DQN) \citep{dabney2018distributional}, the implicit quantile network (IQN) \citep{dabney2018implicit}, and non-crossing quantile regression \citep{zhou2020noncrossing}. Their empirical success extends to high-altitude balloon navigation \citep{bellemare2020autonomous}, robotic manipulation \citep{bodnar2019quantile}, and Atari games \citep{dabney2018distributional,dabney2018implicit}.

The statistical analysis of a quantile representation must account for both the sample size and the number of quantiles. Increasing the number of atoms improves the representation of a return law, but it also brings more parameters and makes the outermost quantile levels closer to zero or one. A small density at such a level can amplify the estimation error of an individual coordinate. In a Wasserstein distance, however, each coordinate has probability mass $1/m$. Whether this averaging compensates for the instability of extreme quantiles is not determined by a coordinatewise analysis.

\citet{cheng2026statistical} developed a model-based analysis of the quantile Bellman fixed point, including a non-asymptotic supremum $W_\infty$ bound with coefficient of order $\sqrt m$ up to logarithmic factors, fixed-resolution asymptotic normality and efficiency, and a limiting covariance analysis. 
These results leave a more specific question: which dependence on the resolution is intrinsic to estimation, and which dependence comes from the choice of loss or from bounding all quantile coordinates by the worst one? In this paper, we answer this question by studying estimation of the projected fixed point under the full family of Wasserstein distances.


The role of the density assumptions is critical. Under our standing assumption that reward densities are continuous on $[0,1]$ and strictly positive in its interior, a uniform lower density bound can fail only at the support endpoints. This motivates two representative regimes: uniformly positive reward densities and densities that vanish at both endpoints. In the latter regime, we impose only monotonicity near the endpoints, without any quantitative restriction on the rate of boundary decay. We show that the effect of endpoint vanishing depends on the Wasserstein order: resolution-independent root-$n$ estimation persists through $W_2$, whereas for higher orders, polynomial dependence on the quantile resolution is unavoidable.

From a theoretical perspective, the quantile-projected Bellman equation couples the quantile coordinates through the continuation distributions, so changing one quantile can affect the Bellman targets at other states. This coupling becomes particularly delicate when reward densities vanish at their endpoints, as extreme quantiles may be sensitive to small perturbations. Although the projected Bellman operator is globally contractive in the supremum $W_\infty$ metric, it is not globally contractive in supremum $W_p$ metric for $p<\infty$. In particular, the supremum $W_\infty$ contraction does not capture the averaging over quantile levels needed for an $m$-independent supremum $W_2$ bound.

\subsection{Our Contributions}
In this paper, we focus on the problem of distributional policy evaluation with quantile parametrization, where the goal is to estimate the return distribution of a given policy using a finite-dimensional quantile representation. 
We consider a $\gamma$-discounted infinite-horizon Markov decision process (MDP) in the tabular setting, where the state space $\gS$ and the action space $\gA$ are finite. Given a policy $\pi$ and quantile level $m$, let $\bm\eta_m$ denote the unique fixed point of the projected distributional Bellman equation $\bm\eta=\bPi_m\gT^\pi \bm\eta$. Here, $\bPi_m$ represents the quantile projection operator and $\gT^\pi$ represents the distributional Bellman operator.

Our goal is to estimate $\bm\eta_m$ when the underlying MDP is unknown. We assume that both the distribution of the random reward and the transition probability of the MDP are unknown. Given an offline dataset consisting of $n|\gS||\gA|$ samples collected from a generative model, we construct the empirical joint distribution $\what Q_{s,a}^{(n)}$ of the reward and next state for each $(s,a)\in\gS\times\gA$. Following the certainty-equivalence principle~\citep{simon1956dynamic,theil1957note}, we estimate $\bm\eta_m$ by the fixed point $\bm\eta_m^{(n)}$ of the empirical projected distributional Bellman operator $\bPi_m \what{\gT}_n^\pi$, where $\what{\gT}_n^\pi$ is constructed from $\{\what Q_{s,a}^{(n)}:(s,a)\in\gS\times\gA\}$. We refer to $\bm\eta_m^{(n)}$ as the quantile fixed point estimator.

Our first contribution is a family of non-asymptotic bounds under supremum $p$-Wasserstein metrics. Under the assumption that the reward density is uniformly lower bounded, Theorem~\ref{thm:A3} gives a bound of order $n^{-1/2}$ simultaneously over every $m$ and every $1\le p\le\infty$. Under the assumption that the reward densities vanish and are monotone at the boundary, Theorem~\ref{thm:A4-W2} gives the same rate for $1\le p\le2$. 
For $p>2$, Corollary~\ref{cor:A4-rootn} gives the rate $O(m^{1/2-1/p}n^{-1/2})$ and an alternative $O(n^{-1/p})$ bound independent of $m$ for every finite $p>2$. Theorem~\ref{thm:A4-Winf} complements these rate statements with an $m$-free $W_\infty$ consistency bound governed by the modulus of continuity of the reward quantile function.

Our second contribution is an exact local statistical benchmark. At each fixed level of quantiles $m$, Theorem~\ref{thm:local-exact} identifies the local asymptotic minimax risk and shows that the quantile fixed point estimator achieves it. Thus, a Gaussian limit distribution characterizes the optimal first-order asymptotic risk of the problem. 

Our third contribution is to establish the sharpness of the resolution dependence through fixed MDP instances. Proposition~\ref{prop:uniform-instance} gives a fully specified MDP with reward density uniformly away from zero, for which the local asymptotic minimax risk constants are bounded above and away from zero uniformly over $m$ and $p$. Theorem~\ref{thm:complexity-hard} constructs an MDP with monotone and vanishing reward density at the boundary for which the local asymptotic minimax risk constants grow as $m^{1/2-1/p}$ up to logarithmic factors along a sequence of $m$ for every $p>2$. Thus, the polynomial resolution exponent in Corollary~\ref{cor:A4-rootn} cannot be reduced in general. 

The key step in the non-asymptotic analysis is to establish a local contraction of the projected Bellman operator in the supremum $W_2$ metric. Combined with careful control of the empirical error, this allows us to obtain an $m$-independent bound even when reward densities vanish at their endpoints. The local minimax result is established separately through a uniform linearization over shrinking Hellinger neighborhoods.

\subsection{Related Work} 
\paragraph{Distributional reinforcement learning}
Distributional reinforcement learning has found applications in areas such as communication network resource management \citep{hua2019gan} and transmission scheduling in vehicular networks \citep{naeem2020generative}. Various approaches have been developed to represent return distributions, including categorical \citep{bellemare2017distributional}, quantile \citep{dabney2018distributional}, and generative-model-based representations \citep{freirich2019distributional}. Among these, quantile-based methods have attracted considerable attention, with developments including quantile regression \citep{dabney2018distributional}, implicit quantile networks \citep{dabney2018implicit}, fully parameterized quantile functions \citep{yang2019fully}, and non-crossing quantile regression \citep{zhou2020noncrossing}.

Theoretical studies have established convergence and statistical guarantees for various distributional reinforcement learning methods. For categorical representations, \citet{rowland2018analysis} analyzed the convergence of categorical temporal-difference learning, while \citet{peng2024statistical} studied its statistical efficiency and \citet{peng2025finite} established finite-sample guarantees under linear function approximation. \citet{wu2023distributional} developed a likelihood-based approach to offline distributional policy evaluation. For quantile temporal-difference learning (QTD), \citet{rowland2023analysis} established convergence results, and \citet{rowland2023statistical} investigated its statistical benefits for expected-value estimation. More recently, \citet{cheng2026qtdfinite} established finite-sample convergence rates for synchronous QTD.

\paragraph{Statistical analysis of reinforcement learning}
Statistical inference for value functions and off-policy evaluation includes high-confidence procedures \citep{thomas2015high,jiang2016doubly}, bootstrap methods \citep{hao2021bootstrapping}, and inference in infinite-horizon settings \citep{shi2022statistical}. For return distributions, existing work has studied CDF estimation \citep{chandak2021universal,huang2022off} and distributional off-policy evaluation \citep{Qi03072025}. Under a generative model, \citet{zhang2025estimation} established finite-sample bounds and asymptotic inference results for return distributions. For quantile distributional reinforcement learning, \citet{cheng2026statistical} established asymptotic normality and semiparametric efficiency of quantile Bellman fixed-point estimators, while \citet{cheng2026online} developed online inference methods with asymptotically valid and anytime-valid confidence intervals.

A related line of work investigates minimax optimality in policy evaluation, for both value functions and return distributions. \citet{pananjady2021instance} established instance-dependent $\ell_\infty$ bounds and minimax guarantees, while \citet{khamaru2021optimal} studied local minimax lower bounds and the optimality of temporal-difference learning. For distributional policy evaluation, \citet{rowland2024nearminimax} established near-minimax sample complexity bounds for a model-based categorical estimator under the supremum $W_1$ metric. Compared with these works, we focus on the quantile Bellman fixed point, characterizing its exact local asymptotic minimax risk under supremum Wasserstein distances and examining how this risk depends on the quantile resolution.

The remainder of this paper is organized as follows. In Section~\ref{sec:prelim}, we introduce some basic concepts of distributional reinforcement learning. In Section~\ref{sec:results}, we present the quantile fixed point estimator and our main theoretical results. In Section~\ref{sec:strategy}, we provide proofs of the results in Section~\ref{sec:results}. We conclude the paper in Section~\ref{sec:conclusion}. Detailed technical ingredients of the proof are given in the appendices.

%% file: sections/preliminaries.tex
In this section, we introduce the necessary background for our work. We review the Markov decision process in Section~\ref{subsec:problem_setting}, followed by metrics on the space of measures in Section~\ref{subsec:measure_metric}. We introduce distributional Bellman operator and quantile projection operator in Section~\ref{subsec:dist_bellman_quantile_projection}. Finally, in Section~\ref{subsec:assumptions}, we present the assumptions under which our main theoretical results are established.

\subsection{Problem Setup}\label{subsec:problem_setting}

We consider a discounted Markov decision process (MDP) specified by 
the tuple $\gM=\<\gS,\gA,\gP_R,P,\gamma\>$, where $\gS$ and $\gA$ are finite state space and finite 
action space respectively, $\gP_R\colon\gS\times\gA\to\Delta([0,1])$ is the distribution of rewards, $P\colon\gS\times\gA\to\Delta(\gS)$ is the transition probability, and $\gamma\in(0,1)$ is the discount factor.
Here $\Delta(\cdot)$ denotes the set of probability distributions over some set.

For a fixed policy $\pi\colon\gS\to\Delta(\gA)$ and an initial state $S_0= s\in\gS$, a random trajectory $\{(S_t,A_t,R_t)\}_{t=0}^\infty$ can be sampled from the MDP using the following procedure: 
\begin{equation*}
    \begin{aligned}
        A_t\mid S_t&\sim\pi(\cdot\mid S_t),\\
        R_t\mid (S_t,A_t)&\sim \gP_R({\cdot}\mid S_t,A_t),\\
        {S_{t+1}}\mid{(S_t,A_t)}&\sim P({\cdot}\mid{S_t,A_t}).\\
    \end{aligned}
\end{equation*}

Here the reward and next state are conditionally independent given $(S_t,A_t)$, as in the model-based formulation.
The return of such a trajectory starting from state $s$ is 
defined as the random variable
\begin{equation*}
    G^\pi(s)\coloneq \sum_{t=0}^\infty \gamma^t R_t,
\end{equation*}
which takes values in $[0,(1-\gamma)^-1]$ almost surely. The value function $V^\pi(s)$ is defined by the expected return $\EB[G^\pi(s)]$. 
We further denote by $\eta^\pi(s) \in \Delta([0, (1-\gamma)^{-1}])$ 
the distribution of $G^\pi(s)$, and denote $\bm\eta^\pi = 
(\eta^\pi(s))_{s \in \gS}$ for the collection of 
return distributions across all states.

\subsection{Metrics on the Space of Measures}\label{subsec:measure_metric}
Denote the space of all probability distributions on $\RB$ as $\sP$. For $\mu\in\sP$, the cumulative distribution function is defined as $F_\mu(x)=\mu(-\infty,x]$, and the quantile function is defined as $F^{-1}_{\mu}(\tau)=\inf\{x\colon F_\mu(x)\geq \tau\}$ for $0<\tau<1$. For $\mu$ with bounded support, we extend its quantile
function to $\tau\in[0,1]$ by continuity. For $1\leq p<\infty$ and $\mu,\nu\in\sP$, the $p$-Wasserstein metric between $\mu$ and $\nu$ is defined as 
\begin{equation*}
    W_p(\mu, \nu)=\left( \int_0^1 \left|F^{-1}_\mu(t)-F^{-1}_\nu(t)\right|^p\mathrm{d}t\right)^{1 / p}, 
\end{equation*}
and the $\infty$-Wasserstein metric is defined as
\begin{equation*}
    W_\infty(\mu, \nu)=\operatorname*{ess\,sup}_{0<t<1} \left|F^{-1}_\mu(t)-F^{-1}_\nu(t)\right|. 
\end{equation*}

Suppose $\mu$ and $\nu$ have cumulative distribution functions $F_\mu$ and $F_\nu$, respectively. 
In the case of $p=1$ we have
\begin{equation*}
    W_1(\mu, \nu)=\int_\RB |F_\mu(x)-F_\nu(x)| \mathrm{d}x.
\end{equation*}
The Kolmogorov--Smirnov metric (KS metric) is defined as
\begin{equation*}
    {\KS}(\mu,\nu)=\sup_{x\in\RB}\abs{F_\mu(x)-F_\nu(x)}.
\end{equation*}

Moreover, for any extended metric $d\colon\sP\times\sP\to[0,\infty]$, we can define its supremum extension $\Bar{d}\colon \sP^\gS\times\sP^\gS\to[0,\infty]$ as
\[
\Bar{d}(\bm\eta,\bm\eta^\prime)=\sup_{s\in\gS}d(\eta(s),\eta^\prime(s)), 
\]
which is an extended metric on $\sP^\gS$.

\subsection{Distributional Bellman Operator and Quantile Projection Operator}\label{subsec:dist_bellman_quantile_projection}

A fundamental property of the value function is that it satisfies 
the Bellman equation. Letting $\bm V^\pi \coloneq (V^\pi(s))_{s\in\gS}$, we have for any $s\in\gS$, 
\begin{equation*}
    \begin{aligned}
            V^\pi(s)&=\brk{T^\pi\bm V^\pi}(s)\\
    &\coloneq \EB_{A\sim\pi(\cdot\mid s), R\sim\gP_R(\cdot\mid s,A)}[R]+\gamma\EB_{A\sim\pi(\cdot\mid s),S^\prime\sim P(\cdot\mid s,A)} [V^\pi(S^\prime)]\\
    &=\sum_{a\in\gA}\pi(a\mid s)\int_0^1 r \gP_R(\mathrm{d}r\mid s,a)+\gamma\sum_{a\in\gA,s^\prime\in\gS} \pi(a\mid s)P(s^\prime\mid s,a)V^\pi(s^\prime).
    \end{aligned}
\end{equation*}
The operator $T^\pi\colon \RB^{\gS}\to \RB^{\gS}$ is referred to as 
the Bellman operator, and the Bellman equation characterizes $\bm V^\pi$ as its unique fixed point.

An analogous relationship holds for the return distributions $\bm\eta^\pi$, known as the distributional Bellman equation. That is, for each $s\in\gS$, 
\begin{equation*}
\begin{aligned}
        \eta^\pi(s)&=\brk{\gT^\pi\bm\eta^\pi}(s)\\
    &\coloneq \EB_{A\sim\pi(\cdot\mid s), R\sim\gP_R(\cdot \mid s,A),S^\prime\sim P(\cdot\mid s,A)}\brk{\prn{b_{R,\gamma}}_\#\eta^\pi(S^\prime)}\\
    &=\sum_{a\in\gA,s^\prime\in\gS}\pi(a\mid s)P(s^\prime\mid s,a)\int_0^1 \prn{b_{r,\gamma}}_\#\eta^\pi(s^\prime)\gP_R(\mathrm{d}r\mid s,a).
\end{aligned}
\end{equation*}
Here $b_{r,\gamma}\colon \RB\to\RB$ denotes the affine map $b_{r,\gamma}(x)=r+\gamma x$, and $g_\#\mu$ is the pushforward of a measure $\mu$ under $g$, defined by $g_\#\mu(B) = \mu(g^{-1}(B))$ for all Borel sets $B$. 
The integral ${\int_0^1 \prn{b_{r,\gamma}}_\#\eta^\pi(s^\prime)\gP_R(\mathrm{d}r\mid s,a)}$ is defined in the sense that for any Borel set $B$,
\begin{equation*}
    \brk{\int_0^1 \prn{b_{r,\gamma}}_\#\eta^\pi(s^\prime)\gP_R(\mathrm{d}r\mid s,a)}(B)=\int_0^1 \brk{\prn{b_{r,\gamma}}_\#\eta^\pi(s^\prime)}(B)\gP_R(\mathrm{d}r\mid s,a). 
\end{equation*}
Recall that $\sP$ is the space of all probability measures on $\RB$, and $\gT^\pi\colon \sP^\gS\to \sP^\gS$ is referred to as the distributional Bellman operator, of which the fixed point is the return distribution $\bm\eta^\pi$.

Because the exact distribution $\bm\eta^\pi$ is infinite-dimensional and cannot be computed exactly, we approximate it using a quantile-parameterized distribution. For $m\in\NB$, the space of all quantile-parametrized probability distributions is defined as
\begin{equation*}
    \sP_{m} \coloneq  \left\{ \nu_{\bx}=\frac{1}{m}\sum_{i=1}^m \delta_{x_i} \mid  \bx=\prn{x_1, \ldots, x_m}^{\top}\in \RB^{m}, x_1\leq \ldots\leq x_m \right\}, 
\end{equation*}
Thus, $\sP_m$ consists of equally weighted mixtures of $m$ Dirac measures.  
We define the quantile projection operator $\Pi_{m}\colon\sP\to\sP_m$ as
\begin{equation*}
   \Pi_{m}\nu=\frac{1}{m}\sum_{i=1}^m\delta_{F^{-1}_{\nu}(\tau_i)}, 
\end{equation*}
where $\tau_i=\frac{2i-1}{2m}$. We lift $\Pi_m$ to the product space $\sP^\gS$ by defining
$(\bPi_m{\bm\eta})(s) \coloneq  \Pi_m\eta(s)$ for any $\bm\eta=(\eta(s))_{s\in\gS}\in\sP^\gS$. 
The properties of $\gT^\pi$ and $\bPi_m$ are summarized in the following proposition. 
\begin{proposition}\cite{bdr2022}\label{prop:basic_properties}
The following statements hold:
\begin{itemize}
    \item $\gT^\pi$ is  $\gamma$-contractive under the $\Wbar_p$ metric for every $p\in[1,\infty]$, namely for every $\bm\eta,\bm\eta^\prime\in\sP^\gS$, 
    \[
    \Wbar_p(\gT^\pi\bm\eta,\gT^\pi\bm\eta^\prime)\leq\gamma\Wbar_p(\bm\eta,\bm\eta^\prime); 
    \]
    \item $\bPi_m$ is non-expansive under the $\Wbar_\infty$ metric, namely $\Wbar_\infty(\bPi_m\bm\eta,\bPi_m\bm\eta^\prime)\leq \Wbar_\infty(\bm\eta,\bm\eta^\prime)$ for every $\bm\eta,\bm\eta^\prime\in\sP^\gS$.  
\end{itemize}
\end{proposition}

It immediately follows from Proposition~\ref{prop:basic_properties} that the quantile projected Bellman operator $\bPi_{m}{\gT}^{\pi}$ 
is a $\gamma$-contraction on the complete space of ordered $m$-atomic distributions supported in $[0,(1-\gamma)^{-1}]$, equipped with $\Wbar_\infty$.
Hence, the quantile projected Bellman equation $\bm\eta=\bPi_{m}\gT^{\pi}\bm\eta$ admits a unique solution $\bm\eta_{m}$. 
We define the quantile parameter $\bm\theta_m\in\RB^{\gS\times [m]}$ by 
\begin{equation*}
    \eta_m(s)=\frac{1}{m}\sum_{i=1}^m\delta_{\theta_m(s,i)},
\end{equation*}
with $\theta_m(s,1)\leq \cdots \leq\theta_m(s,m)$ for every $s\in\gS$, and $[m]=\{1,2,\ldots,m\}$.

Write $B=(1-\gamma)^{-1}$ and $I=[0,B]$. We define
\[
\Theta_m=\left\{\btheta\in I^{\mathcal S\times[m]}:\theta(s,1)\le\cdots\le\theta(s,m)\text{ for every }s\in\mathcal S\right\}.
\]
For $\btheta\in\Theta_m$, write
\[
\eta_\btheta(s)=\frac1m\sum_{i=1}^m\delta_{\theta(s,i)},\qquad\bm\eta_\btheta=(\eta_\btheta(s))_{s\in\mathcal S}.
\]
For $\bz\in\mathbb R^{\mathcal S\times[m]}$, define
\[
\ell_{m,p}(\bz)=\max_{s\in\mathcal S}\left(\frac1m\sum_{i=1}^m|z(s,i)|^p\right)^{1/p},\quad 1\le p<\infty,
\]
and
\[
\ell_{m,\infty}(\bz)=\max_{s,i}|z(s,i)|.
\]
The quantile representation gives
\begin{equation}\label{eq:Wp-array}
    \overline W_p(\bm\eta_\btheta,\bm\eta_{\bm\vartheta})=\ell_{m,p}(\btheta-\bm\vartheta),\qquad 1\le p\le\infty.
\end{equation}

\subsection{Main Assumptions}\label{subsec:assumptions}
For every $(s,a)\in\gS\times\gA$, we make the following assumption about $\gP_R(\cdot\mid s,a)$. 
\begin{assumption}[Positive reward densities]\label{ass:A1}
For any $s\in\gS$ and $a\in\gA$, $\gP_R(\cdot\mid s,a)$ is supported on $[0,1]$ and has a Lebesgue density $p_{s,a}$. Moreover, $p_{s,a}$ is continuous on $[0,1]$, and there is a positive constant $C_0$ such that $0<p_{s,a}(x)\le C_0$ for every $x\in(0,1)$.
\end{assumption}
The following proposition, proved in~\citet{cheng2026statistical}, shows that this assumption gives a well-behaved return density. 
\begin{proposition}\label{prop:return_density}
For every $s\in\gS$, $\eta^\pi(s)$ has a continuous Lebesgue density $p_{\eta^\pi(s)}$ upper bounded by $C_0$ on $[0,(1-\gamma)^{-1}]$. Moreover, $p_{\eta^\pi(s)}(x)>0$ for every $x\in(0,(1-\gamma)^{-1})$, and $p_{\eta^\pi(s)}(0)=p_{\eta^\pi(s)}((1-\gamma)^{-1})=0$.
\end{proposition}

We also impose a non-boundary condition on the population quantile parameters.
\begin{assumption}[No boundary coincidence]\label{ass:A2}
For every $m\ge1$ and $(s,i),(s^\prime,j)\in\gS\times[m]$,
\[
 \theta_m(s,i)-\gamma\theta_m(s^\prime,j)\notin\{0,1\}.
\]
\end{assumption}
This condition excludes configurations in which a Bellman-shifted quantile coincides with an endpoint of the reward support. Together with Assumption~\ref{ass:A1}, it gives positive target density at each population midpoint quantile and differentiability of the fixed-point CDF equation near the true parameter. We suppose that assumptions~\ref{ass:A1} and~\ref{ass:A2} hold throughout the paper.

For the non-asymptotic analysis we impose one of two alternative conditions. The first requires a positive lower reward density, as in the usual nondegenerate quantile-estimation setting.
\begin{assumption}[Uniform lower density]\label{ass:A3}
There is a constant $c_0>0$ such that $p_{s,a}(x)\ge c_0$ for every $s\in\gS$, $a\in\gA$, and $x\in[0,1]$.
\end{assumption}
The second allows the reward density to vanish at the support endpoints, replacing the lower bound by a qualitative shape condition.
\begin{assumption}[Regular endpoint vanishing]\label{ass:A4}
For every $s\in\gS$ and $a\in\gA$, $p_{s,a}(0)=p_{s,a}(1)=0$. There is a common $\kappa>0$ such that each $p_{s,a}$ is nondecreasing on $[0,\kappa]$ and nonincreasing on $[1-\kappa,1]$.
\end{assumption}
The endpoint monotonicity excludes oscillatory boundary behaviour but places no quantitative restriction on the speed at which the density vanishes. Beta densities with both shape parameters greater than one are examples. The two assumptions describe complementary density regimes.

Under Assumption~\ref{ass:A4}, zero extension makes each reward density continuous on $\RB$. Since there are finitely many state--action pairs, these extensions have the common modulus of continuity
\begin{equation*}
 \varpi_R(h)\coloneq\max_{s,a}\sup_{\substack{x,y\in\RB\\|x-y|\le h}}
 |p_{s,a}(x)-p_{s,a}(y)|,
 \qquad \lim_{h\downarrow0}\varpi_R(h)=0.
\end{equation*}


\paragraph{Notation and constants}
Unless otherwise specified, $C,c,C',\ldots$ denote positive universal constants, while $C_\gM,c_\gM$ denote positive constants depending only on the fixed MDP, including the fixed policy, but not on $m,n,p$, or $\delta$. Additional dependencies are indicated explicitly, as in $C_{m,c,Q_0}$. The values of constants may change between occurrences. We write $a\asymp b$ if $cb\le a\le Cb$ for positive constants uniform over the ranges under consideration, with dependencies understood as above.


%% file: sections/results.tex
In this section, we first introduce the quantile fixed point estimator $\bm\eta_m^{(n)}$ and establish its non-asymptotic Wasserstein convergence rates under the two reward-density regimes. We then characterize the exact local asymptotic minimax risk and use two fixed MDPs to examine the sharpness of the resulting sample-size and resolution dependence.

\subsection{The Quantile Fixed Point Estimator}\label{subsec:estimator}
We assume access to a generative model for the unknown MDP. For each $(s,a)\in\gS\times\gA$, we observe
\[
 X_{k}^{(s,a)}=(R_k^{(s,a)},S_k^{\prime(s,a)})\simiid Q_{0,s,a}\coloneq \gP_R(\cdot\mid s,a)\otimes P(\cdot\mid s,a),
 \ k\in[n].
\]
The samples are independent across different $(s,a)$. For subsequent analysis, we write $Q_0=(Q_{0,s,a})_{(s,a)\in\gS\times\gA}$.


Given the samples, we construct the empirical distributions
\begin{equation*}
 \what Q_{s,a}^{(n)}=\frac1n\sum_{k=1}^n
 \delta_{(R_k^{(s,a)},S_k^{\prime(s,a)})}
\end{equation*}
and define the empirical distributional Bellman operator by
\begin{equation*}
    [\widehat \gT_n^\pi\bm\eta](s)=\sum_{a\in\gA}\pi(a\mid s)\int (b_{r,\gamma})_\#\eta(s')\,\widehat Q_{s,a}^{(n)}(\rd(r,s')).
\end{equation*}
Then the empirical quantile projected Bellman equation
$\bm\eta=\bPi_{m}\what{\gT}^\pi_n\bm\eta$
admits a unique solution $\bm\eta^{(n)}_{m}$. We also define $\bm\theta^{(n)}_m\in\RB^{\gS\times [m]}$ by
\begin{equation*}
\eta^{(n)}_m(s)=\frac{1}{m}\sum_{i=1}^m\delta_{\theta^{(n)}_m(s,i)},
\end{equation*}
with $\theta^{(n)}_m(s,1)\leq\ldots\leq\theta^{(n)}_m(s,m)$ for every $s\in\gS$. We call $\bm\eta_m^{(n)}$ the quantile fixed point estimator throughout the paper.

\subsection{Uniformly Positive Reward Densities}\label{subsec:A3-results}
We first consider Assumption~\ref{ass:A3}. The next theorem shows that increasing the quantile resolution does not worsen the root-$n$ estimation order, even in the strongest Wasserstein metric.

\begin{theorem}\label{thm:A3}
Under Assumption~\ref{ass:A3}, there exist instance-dependent constants $C_1(\gM),C_2(\gM)\in(0,\infty)$ such that for every $\delta\in(0,1)$ and every $n\ge1$,
\begin{equation}\label{eq:A3-hp}
  \PB\left\{\sup_{m\ge1}\Wbar_\infty(\Beta_m^{(n)},\Beta_m)\le C_1(\gM)\sqrt{\frac{\log(4|\gS|/\delta)}n}\right\}\ge1-\delta.
\end{equation}
Moreover,
\[
  \E\sup_{m\ge1}\Wbar_\infty(\Beta_m^{(n)},\Beta_m)
  \le \frac{C_2(\gM)}{\sqrt n}.
\]
\end{theorem}

Since $\Wbar_p\le\Wbar_\infty$ for every $1\le p<\infty$, both bounds in Theorem~\ref{thm:A3} also hold with $\Wbar_\infty$ replaced by $\Wbar_p$, simultaneously for all such $p$.
The proof separates the errors in estimating reward distributions and transition probabilities, and controls the latter through relative deviations before applying the global $\Wbar_\infty$ contraction. This avoids bounding all quantile errors by the sensitivity of the extreme quantiles. The complete proof is in Section~\ref{subsec:proof-A3}.

\subsection{Endpoint-Vanishing Reward Densities}\label{subsec:A4-results}
Assumption~\ref{ass:A4} removes the lower density bound and allows arbitrary boundary decay subject to monotonicity. The root-$n$ rate remains independent of $m$ under $\overline{W}_2$ metric.

\begin{theorem}\label{thm:A4-W2}
Under Assumption~\ref{ass:A4}, there exist instance-dependent constants $C_1(\gM),C_2(\gM)\in(0,\infty)$ such that for every $n\ge1$ and $\delta\in(0,1)$,
\begin{equation}\label{eq:A4-W2-hp}
 \PB\left\{\sup_{m\ge1}\Wbar_2(\Beta_m^{(n)},\Beta_m)\le C_1(\gM)\sqrt{\frac{\log(4|\gS|/\delta)}n}\right\}\ge1-\delta.
\end{equation}
Moreover,
\begin{equation*}
 \E\sup_{m\ge1}\Wbar_2(\Beta_m^{(n)},\Beta_m)\le\frac{C_2(\gM)}{\sqrt n}.
\end{equation*}
\end{theorem}

Since $\Wbar_p\le\Wbar_2$ for every $1\le p\le2$, both bounds also hold for all such $p$. The difficulty here is that the projected Bellman operator $\bPi_m\gT^\pi$ need not be globally contractive under the $\Wbar_2$ metric. We prove that it is contractive on a common neighborhood of $\bm\eta^\pi$ for all sufficiently large $m$; the finitely many smaller resolutions are handled separately. Section~\ref{subsec:proof-A4} gives the proof.
We have the following corollary from Theorem~\ref{thm:A4-W2}. 
\begin{corollary}\label{cor:A4-rootn}
Under the assumptions of Theorem~\ref{thm:A4-W2}, for every $n\ge1$ and $\delta\in(0,1)$, on the event in \eqref{eq:A4-W2-hp}, which has probability at least $1-\delta$, the following bounds hold simultaneously for every $m\ge1$ and every $2\le p\le\infty$:
\begin{align}
 \Wbar_p(\Beta_m^{(n)},\Beta_m)
 &\le C_1(\gM)m^{1/2-1/p} \sqrt{\frac{\log(4|\gS|/\delta)}n}\label{eq:larger-p-confidence}\\
 \sup_{m\ge1}\Wbar_p(\Beta_m^{(n)},\Beta_m)
 &\le C_1(\gM)^{2/p}(1-\gamma)^{-(1-2/p)}\prn{\frac{\log(4|\gS|/\delta)}n}^{1/p}\label{eq:larger-p-mfree}
\end{align}
where $1/\infty=0$ and $C_1(\gM)$ is the constant defined in Theorem~\ref{thm:A4-W2}. 
\end{corollary}

Equation~\eqref{eq:larger-p-mfree} in Corollary~\ref{cor:A4-rootn} does not give consistency in $W_\infty$. However, we can still prove an $m$-free $W_\infty$ consistency bound under Assumption~\ref{ass:A4}, at a rate determined by the continuity modulus of the reward quantile functions. Write $F_{s,a}$ for the CDF of $\gP_R(\cdot\mid s,a)$ and define
\begin{equation*}
 \omega_R(h)\coloneq \max_{s,a}\sup_{\substack{u,v\in[0,1]\\|u-v|\le h}}
 \left|F_{s,a}^{-1}(u)-F_{s,a}^{-1}(v)\right|,
 \qquad 0\le h\le1.
\end{equation*}
By Assumption~\ref{ass:A1}, $\omega_R(h)\to0$ as $h\downarrow0$.
\begin{theorem}\label{thm:A4-Winf}
Under Assumption~\ref{ass:A4}, there is an instance-dependent constant $K_3(\gM)\in(0,\infty)$ such that, for every $n\ge1$ and $\delta\in(0,1)$, with
\begin{equation*}
 h_{n,\delta}\coloneq\min\left\{1,\frac{16}{p_*}\sqrt{\frac{\log(4|\gS|/\delta)}{n}}\right\},
\end{equation*}
where
\begin{equation*}
    p_*=\min_{(s,a,s^\prime):P(s^\prime\mid s,a)>0}P(s^\prime\mid s,a),
\end{equation*}
we have
\begin{equation*}
 \PB\left\{\sup_{m\ge1}\Wbar_\infty(\Beta_m^{(n)},\Beta_m)\le K_3(\gM)\bigl[\omega_R(h_{n,\delta})+h_{n,\delta}\bigr]\right\}\ge1-\delta.
\end{equation*}
\end{theorem}

\subsection{Local Asymptotic Minimaxity}\label{sec:local}\label{subsec:local-model}
The preceding results provide upper bounds for the quantile fixed point estimator. We now characterize the exact local asymptotic minimax risk at each fixed resolution. Consider the model class
\begin{equation*}
 \gQ=\left\{Q=(Q_{s,a})_{(s,a)\in\gS\times\gA}:Q_{s,a}\in\Delta([0,1]\times\gS),\quad
 Q_{s,a}\ll\lambda\otimes\#_\gS\right\},
\end{equation*}
where $\lambda$ is Lebesgue measure on $[0,1]$ and $\#_\gS$ is counting measure on $\gS$. For any $Q\in\gQ$, define the distributional Bellman operator
\[
 [\gT_Q^\pi\bm\eta](s)=\sum_{a\in\gA}\pi(a\mid s)\int (b_{r,\gamma})_\#\eta(s')\,Q_{s,a}(d(r,s')).
\]
Let $\Beta_m(Q)$ denote the unique fixed point of $\bPi_m\gT_Q^\pi$, with ordered quantile parameter $\bm\theta_m(Q)$. In particular, $\Beta_m(Q_0)=\Beta_m$ and $\bm\theta_m(Q_0)=\bm\theta_m$. Under $Q$, the sampling experiment is $\PB_Q^{(n)}=\bigotimes_{s,a}Q_{s,a}^{\otimes n}$, and the estimation target is $\Beta_m(Q)$.

For $Q,Q_0\in\gQ$, define the block Hellinger distance and the local neighborhood by
\begin{equation*}
 d_H(Q,Q_0)^2=\sum_{s,a}\int\bigl(\sqrt{\rd Q_{s,a}}-\sqrt{\rd Q_{0,s,a}}\bigr)^2,\ \gN_n(c;Q_0)=\{Q\in\gQ:d_H(Q,Q_0)\le c/\sqrt n\}.
\end{equation*}
For fixed $m$ and $1\le p\le\infty$, the standardized local minimax risk is defined as
\begin{equation*}
 r_{n,m,p}(c;Q_0)=\inf_{\what\Beta_n}\sup_{Q\in\gN_n(c;Q_0)}
 \E_Q\!\left[\sqrt n\,\Wbar_p\bigl(\what\Beta_n,\Beta_m(Q)\bigr)\right], 
\end{equation*}
where the infimum ranges over all possibly randomized measurable estimators taking values in $\Delta(I)^{\gS}$.
We define the lower and upper local asymptotic minimax risks by
\begin{equation*}
 \underline{\gR}^{\mathrm{loc}}_{m,p}(Q_0)=\lim_{c\to\infty}\liminf_{n\to\infty}r_{n,m,p}(c;Q_0),\qquad
 \overline{\gR}^{\mathrm{loc}}_{m,p}(Q_0)=\lim_{c\to\infty}\limsup_{n\to\infty}r_{n,m,p}(c;Q_0).
\end{equation*}
Both outer limits exist by monotonicity in $c$. 

To characterize these risks, we introduce the Gaussian limit associated with the quantile fixed point estimator. All population quantities below are evaluated at $Q_0$. Let $d=|\gS|m$ and write
\[
 y_{m,s,i}(r,s^\prime)=\frac1m\sum_{j=1}^m\ind\{r+\gamma\theta_m(s^\prime,j)\le\theta_m(s,i)\}.
\]
Let $f_{m,s}$ denote the density of $[\gT^\pi\Beta_m](s)$. Define the matrices $\bG_m,\bSigma_m\in\RB^{d\times d}$ by
\begin{equation*}
\begin{aligned}
 [\bG_m]_{(s,i),(s^\prime,j)}&=\ind\{(s,i)=(s^\prime,j)\}f_{m,s}(\theta_m(s,i))\\
 &\quad-\frac{\gamma}{m}\sum_a\pi(a\mid s)P(s^\prime\mid s,a)p_{s,a}\bigl(\theta_m(s,i)-\gamma\theta_m(s^\prime,j)\bigr),
\end{aligned}
\end{equation*}
and
\begin{equation*}
 [\bSigma_m]_{(s,i),(s^\prime,j)}=\ind\{s=s^\prime\}\sum_a\pi(a\mid s)^2\cov_{Q_{0,s,a}}\bigl(y_{m,s,i}(R,S^\prime),y_{m,s,j}(R,S^\prime)\bigr).
\end{equation*}
Then we can characterize the local risk constant in the following theorem. 
\begin{theorem}\label{thm:local-exact}
Fix $m\ge1$ and $1\le p\le\infty$. Denote $\bV_m=\bG_m^{-1}\bSigma_m\bG_m^{-\top}$ and let
$\bZ_m\sim\normal(0,\bV_m)$ and define
\begin{equation*}
\mathfrak C_{m,p}(Q_0)\coloneq\E\ell_{m,p}(\bZ_m).
\end{equation*}
Then
\begin{equation}\label{eq:local-exact}
 \underline{\gR}^{\mathrm{loc}}_{m,p}(Q_0)
 =\overline{\gR}^{\mathrm{loc}}_{m,p}(Q_0)
 =\mathfrak C_{m,p}(Q_0).
\end{equation}
Therefore the quantile fixed point estimator attains this constant. More precisely, for every fixed $c<\infty$,
\begin{equation}\label{eq:local-attainment}
 \lim_{n\to\infty}\sup_{Q\in\gN_n(c;Q_0)}
 \left|\E_Q\!\left[\sqrt n\,\Wbar_p\bigl(\Beta_m^{(n)},\Beta_m(Q)\bigr)\right]
 -\mathfrak C_{m,p}(Q_0)\right|=0.
\end{equation}
\end{theorem}

This theorem identifies the exact local asymptotic minimax risk under expected Wasserstein loss, providing a statistical benchmark for assessing the resolution dependence of the estimation error. The key to the upper bound is to establish uniform linearization and uniform integrability over the full Hellinger neighborhood. For the matching lower bound, we construct a parametric submodel whose local Gaussian experiment yields the same risk.

\subsection{Sharpness of the Wasserstein Rates}\label{sec:resolution}
Theorem~\ref{thm:local-exact} characterizes the optimal local asymptotic risk at each fixed resolution $m$. We now examine its dependence on $m$ to determine whether the resolution factors in our non-asymptotic upper bounds reflect an intrinsic statistical difficulty. We first consider an MDP with a uniformly positive reward density.
\begin{proposition}\label{prop:uniform-instance}
Consider one state, one action, a self-transition, $R\sim\mathrm{Unif}[0,1]$, and $\gamma=e^{-1}$. This fixed MDP satisfies Assumptions~\ref{ass:A1}--\ref{ass:A3} for every $m$. For every $m\geq 1$, $1\leq p\leq\infty$, we have $\mathfrak C_{m,p}\asymp1$. 
\end{proposition}
Together with Theorem~\ref{thm:local-exact}, this proposition shows that the $n^{-1/2}$ order in Theorem~\ref{thm:A3} cannot be improved in general. The next theorem shows that the polynomial dependence on $m$ under Assumption~\ref{ass:A4} can also be intrinsic to the estimation problem.
\begin{theorem}\label{thm:complexity-hard}
There exists a fixed two-state, one-action MDP $Q_*$ with $\gamma=1/2$, transitions $P(t\mid s)=P(t\mid t)=1$, and satisfying Assumptions~\ref{ass:A1},~\ref{ass:A2}, and~\ref{ass:A4}, together with a sequence $m_j\to\infty$, such that
\begin{equation*}
 \frac{c_*\,m_j^{1/2-1/p}}{(\log m_j)^{K_*}}\le\mathfrak C_{m_j,p}(Q_*)\le C_*\,m_j^{1/2-1/p},\ 2<p\le\infty,
\end{equation*}
for all sufficiently large $j$. Here $c_*,C_*,K_*$ are constants determined by the MDP $Q_*$ and are independent of $j,m_j,n,p$. For this same instance $\mathfrak C_{m,p}(Q_*)\asymp1$ for $1\le p\le2$.
\end{theorem}

Theorem~\ref{thm:complexity-hard} shows that the polynomial exponent $1/2-1/p$ in Corollary~\ref{cor:A4-rootn} is optimal in the local asymptotic minimax sense, up to logarithmic factors. Moreover, the same MDP has resolution-independent local risk constants for $1\le p\le2$. Thus, the distinction between lower and higher Wasserstein orders reflects a genuine difference in statistical difficulty.

The key to the construction is to design a fixed reward density with infinitely many regions of very small density near the boundary. We select a sequence of resolutions $m_j$ such that the smallest quantile at state $s$ lies in these regions. At these resolutions, the Bellman target density at the smallest quantile becomes extremely small, making its estimation highly sensitive to sampling fluctuations. Consequently, the asymptotic variance of this coordinate grows nearly linearly with $m_j$. 

%% file: sections/proof_outlines.tex
In this section, we prove the results in Section~\ref{sec:results}. 

\subsection{Analysis under Uniformly Positive Reward Densities}\label{subsec:proof-A3}
The proof of Theorem~\ref{thm:A3} consists of two steps. We first establish Lemma~\ref{lem:A3-one-step}, a concentration inequality for the empirical Bellman operator evaluated at the population fixed point $\bm\eta_m$. To derive an upper bound independent of $m$, we separate the errors in estimating the reward distributions and transition probabilities. The lower density bound controls the former, while relative deviations of the transition probabilities control the latter. The detailed proof is given in Appendix~\ref{app:proof-A3-group}. We then combine this estimate with the $\gamma$-contraction property of $\bPi_m\gT^\pi$ under $\Wbar_\infty$ metric to obtain a fixed-point perturbation bound, which implies Theorem~\ref{thm:A3}. 
\begin{lemma}\label{lem:A3-one-step}
Under Assumption~\ref{ass:A3}, define
\[
 \Delta_n\coloneq \sup_{m\ge1}\Wbar_\infty\prn{\bPi_m\what{\gT}_n^\pi\Beta_m,\bPi_m\gT^\pi\Beta_m}.
\]
There is a constant $K_\gM$ independent of $m$ and $n$ such that, for every $n\ge1$ and $\delta\in(0,1)$,
\[
 \PB\brk{\Delta_n>K_\gM\sqrt{\frac{\log(4|\gS|/\delta)}n}}\le\delta.
\]
\end{lemma}

The contraction of $\bPi_m\what{\gT}_n^\pi$ now transfers this bound to the fixed points. Indeed,
\begin{align*}
 \Wbar_\infty(\Beta_m^{(n)},\Beta_m)
 &\le\Wbar_\infty\prn{\bPi_m\what{\gT}_n^\pi\Beta_m^{(n)},
                         \bPi_m\what{\gT}_n^\pi\Beta_m}+\Delta_n\\
 &\le\gamma\Wbar_\infty(\Beta_m^{(n)},\Beta_m)+\Delta_n.
\end{align*}
Rearranging the inequality above yields Equation~\eqref{eq:A3-hp}, and integrating its tail bound gives the expectation bound. This proves Theorem~\ref{thm:A3}.

\subsection{Analysis under Endpoint-Vanishing Reward Densities}\label{subsec:proof-A4}
When reward densities vanish at the endpoints, a small error in a distribution function can produce a large error in an extreme quantile. To prove Theorem~\ref{thm:A4-W2}, we first show that for sufficiently large $m$, the population and empirical fixed points lie in a common neighborhood of $\bm\eta^\pi$, where the population quantile Bellman operator is contractive in $\Wbar_2$. 
We then establish a concentration bound for the CDFs of empirical Bellman operator evaluated at distributions $\bm\eta_\btheta$, uniformly over $\bm\eta_\btheta$ in this neighborhood. A quantile inversion inequality converts this bound into a Wasserstein error bound, which can then be combined with the local contraction.

First we introduce some notations. For $\btheta\in\Theta_m$, write
\begin{equation*}
 F_{s,\btheta}(x)\coloneq F_{[\gT^\pi\Beta_{\btheta}](s)}(x),
 \qquad
 \what F_{s,\btheta}^{(n)}(x)\coloneq F_{[\what{\gT}_n^\pi\Beta_{\btheta}](s)}(x),
 \qquad F_{m,s}\coloneq F_{s,\btheta_m}.
\end{equation*}
The corresponding maps on quantile parameters are
\begin{equation*}
 [\Phi_m(\btheta)]_{s,i}=F_{s,\btheta}^{-1}(\tau_i),
 \qquad
 [\what\Phi_{m,n}(\btheta)]_{s,i}
   =(\what F_{s,\btheta}^{(n)})^{-1}(\tau_i).
\end{equation*}
In this subsection, write $\|\bz\|_{2,m}=\ell_{m,2}(\bz)$. Under Assumption~\ref{ass:A4}, the density of $F_{s,\btheta}$ is
\begin{equation}\label{eq:target-density}
 f_{s,\btheta}(x)=\sum_{a,s^\prime,j}
 \frac{\pi(a\mid s)P(s^\prime\mid s,a)}m
 p_{s,a}(x-\gamma\theta(s^\prime,j)),
\end{equation}
where reward densities are extended by zero outside $[0,1]$. Define
\begin{equation*}
 \gU_m(\rho)=\{\btheta\in\Theta_m:
          \Wbar_1(\Beta_{\btheta},\Beta^\pi)\le\rho\}.
\end{equation*}
To control the empirical error, let
\[
 J_{s,a,s^\prime}(x)=P(s^\prime\mid s,a)F_{s,a}(x),
 \qquad
 \what J_{s,a,s^\prime}(x)=\frac1n\sum_{k=1}^n
 \ind\{S_k^{\prime(s,a)}=s^\prime,\ R_k^{(s,a)}\le x\},
\]
and set
\begin{equation}\label{eq:Dn-definition}
 D_n=\max_s\sum_a\pi(a\mid s)\sum_{s^\prime}
 \norm{\what J_{s,a,s^\prime}-J_{s,a,s^\prime}}_\infty.
\end{equation}

To establish the $\Wbar_2$ bound, we first restrict the analysis to a neighborhood of the return distribution $\Beta^\pi$. The population fixed point $\Beta_m$ approaches $\Beta^\pi$ as $m$ increases, while the empirical fixed point remains close when the sampling error $D_n$ is small. In this neighborhood, the densities of the Bellman targets are uniformly bounded away from zero in the interior. The following lemma makes these statements precise.

\begin{lemma}\label{lem:common-localization}
Under Assumption~\ref{ass:A4}, fix
$0<\ell<\min\{\kappa,1/2,B/2\}$.
There exist constants $\rho_0=\rho_0(\gM,\ell)>0$ and $c_{\gM,\ell}>0$ such that, for every $m\ge1$ and $\btheta\in\gU_m(\rho_0)$,
\[
 \inf_{s\in\gS}\inf_{x\in[\ell,B-\ell]}
 f_{s,\btheta}(x)\ge c_{\gM,\ell}.
\]
Moreover, for each $0<\rho\le\rho_0$, there exist $\varepsilon_0>0$ and $M_1<\infty$, depending only on $\gM$, $\ell$, and $\rho$,
such that, on $\{D_n\le\varepsilon_0\}$, for $m\geq M_1$,
\[
 \btheta_m,\btheta_m^{(n)}\in\gU_m(\rho).
\]
Finally, for each fixed $M<\infty$, there is a constant $C_{M,\gM}<\infty$ such that
\[
 \max_{1\le m\le M}
 \Wbar_\infty(\Beta_m^{(n)},\Beta_m)
 \le C_{M,\gM}D_n.
\]
\end{lemma}

We next establish a contraction of the population quantile map $\Phi_m$ in the $\|\cdot\|_{2,m}$ norm on this neighborhood.

\begin{lemma}\label{lem:W2-contraction}
Under Assumption~\ref{ass:A4}, there exist $\rho>0$, $M_0<\infty$ and $\lambda_2\in(0,1)$, which only depend on $\gM$, such that for all $m\ge M_0$ and $\btheta,\bvartheta\in\gU_m(\rho)$,
\[
 \|\Phi_m(\btheta)-\Phi_m(\bvartheta)\|_{2,m}
 \le\lambda_2\|\btheta-\bvartheta\|_{2,m}.
\]
\end{lemma}


After establishing the local contraction, it remains to control the difference between the quantiles of $\widehat{\gT}_n^\pi\bm\eta_\btheta$ and $\gT^\pi\bm\eta_\btheta$ for inputs in the common neighborhood. Since the population density may be very small near the endpoints, we need a CDF error bound that takes the density into account. The following lemma, proved in Appendix~\ref{app:proof-square-inverse-additive}, shows how such a bound controls the average squared quantile error uniformly over $m$. 
\begin{lemma}\label{lem:square-inverse-additive}
Under Assumption~\ref{ass:A4}, fix $\ell$ as in Lemma~\ref{lem:common-localization} and let $\rho_0=\rho_0(\gM,\ell)$ be its neighborhood radius. For $n,m\ge1$, $0<\rho\le\rho_0$, $\btheta\in\gU_m(\rho)$ and $s\in\gS$, suppose that
\[
 \abs{\what F_{s,\btheta}^{(n)}(x)-F_{s,\btheta}(x)}
 \le z\sqrt{f_{s,\btheta}(x)}+\varepsilon,
 \qquad x\in I,
\]
for some $z,\varepsilon\ge0$. Then
\begin{equation}\label{eq:additive-inverse-result}
 \frac1m\sum_{i=1}^m
 \abs{[\what\Phi_{m,n}(\btheta)]_{s,i}-[\Phi_m(\btheta)]_{s,i}}^2
 \le C_{\gM,\ell}(z^2+\varepsilon),
\end{equation}
where $C_{\gM,\ell}$ depends only on $\gM$ and $\ell$.
\end{lemma}

Finally, the concentration bound below has exactly the form required by the inversion result above. Its proof is in Appendix~\ref{app:proof-joint-bernstein-event}.
\begin{lemma}\label{lem:joint-bernstein-event}
Under Assumption~\ref{ass:A4}, choose $\rho>0$ sufficiently small so that Lemmas~\ref{lem:common-localization} and~\ref{lem:W2-contraction} apply, and let $\varepsilon_0$ be the corresponding constant in Lemma~\ref{lem:common-localization}. There exist instance-dependent constants $K_\gM,c_\gM^*>0$ such that, for $1\le t\le c_\gM^*n$, an event $\gE_{n,t}$ of probability at least $1-4|\gS|e^{-t}$ satisfies
\[
 D_n\le\varepsilon_0,
 \qquad D_n\le\sqrt2\,|\gS|\sqrt{t/n},
\]
and, simultaneously for all $m\ge1$, $\btheta\in\gU_m(\rho)$, $s\in\gS$, and $x\in[0,B]$,
\begin{equation}\label{eq:joint-bernstein-target}
 \abs{\what F_{s,\btheta}^{(n)}(x)-F_{s,\btheta}(x)}
 \le K_\gM\left\{\sqrt{\frac{t f_{s,\btheta}(x)}n}+\frac tn\right\}.
\end{equation}
\end{lemma}

We now combine these results. Choose $\ell$ and $0<\rho\le\rho_0(\gM,\ell)$, depending only on $\gM$, so that Lemmas~\ref{lem:common-localization},~\ref{lem:W2-contraction} and~\ref{lem:joint-bernstein-event} apply. On $\gE_{n,t}$, both fixed points belong to $\gU_m(\rho)$ for $m\ge M_*\coloneq\max\{M_0,M_1\}$. Applying Lemma~\ref{lem:square-inverse-additive} to~\eqref{eq:joint-bernstein-target} with $z=K_\gM\sqrt{t/n}$ and $\varepsilon=K_\gM t/n$ gives
\[
 \sup_{m\ge1}\sup_{\btheta\in\gU_m(\rho)}
 \norm{\what\Phi_{m,n}(\btheta)-\Phi_m(\btheta)}_{2,m}^2
 \le C_\gM\frac tn.
\]
Since $\what\Phi_{m,n}(\btheta_m^{(n)})=\btheta_m^{(n)}$ and $\Phi_m(\btheta_m)=\btheta_m$, Lemma~\ref{lem:W2-contraction} yields
\[
 \norm{\btheta_m^{(n)}-\btheta_m}_{2,m}
 \le C_\gM\sqrt{\frac tn}
       +\lambda_2\|\btheta_m^{(n)}-\btheta_m\|_{2,m}.
\]
For $m<M_*$, the last assertion of Lemma~\ref{lem:common-localization} gives the same order. Combining the two ranges of $m$, we obtain
\begin{equation}\label{eq:Vn-exp-tail}
 \PB\left\{\sup_{m\ge1}\Wbar_2\prn{\Beta_m^{(n)},\Beta_m}
 >K_0(\gM)\sqrt{\frac tn}\right\}\le4|\gS|e^{-t},
 \qquad t\ge1,
\end{equation}
for an instance-dependent $K_0(\gM)$. For $t>c_\gM^*n$, the support bound gives the result after taking $K_0(\gM)\ge B/\sqrt{c_\gM^*}$. Setting $t=\log(4|\gS|/\delta)$ proves \eqref{eq:A4-W2-hp}, and integrating \eqref{eq:Vn-exp-tail} gives the expectation bound.

For larger Wasserstein orders, Corollary~\ref{cor:A4-rootn} follows from two norm comparisons. For $2\le p\le\infty$,
\[
 \ell_{m,p}(\bz)\le m^{1/2-1/p}\ell_{m,2}(\bz).
\]
Apply this to $\btheta_m^{(n)}-\btheta_m$ and use \eqref{eq:Wp-array} on the event in \eqref{eq:A4-W2-hp} to obtain \eqref{eq:larger-p-confidence}. For $2\le p<\infty$ and distributions $\mu,\nu$ on $I$,
\[
 W_p(\mu,\nu)^p
 =\int_0^1|F_\mu^{-1}(u)-F_\nu^{-1}(u)|^p\,\rd u
 \le B^{p-2}W_2(\mu,\nu)^2.
\]
Applying this inequality at each state on the same event gives \eqref{eq:larger-p-mfree}.

To prove Theorem~\ref{thm:A4-Winf}, we first bound the error in one empirical Bellman update at the population fixed point $\Beta_m$. We separate the errors in estimating the reward distributions and transition probabilities, and use the continuity modulus of the reward quantile functions to control the former. The resulting bound is given in the following lemma, proved in Appendix~\ref{app:proof-A4-Winf-one-step}.
\begin{lemma}\label{lem:A4-Winf-one-step}
Under Assumption~\ref{ass:A4}, there is a constant $K_\gM$ only depending on $\gM$ such that, for every $n\ge1$ and $\delta\in(0,1)$,
\begin{equation}\label{eq:A4-Winf-one-step}
 \PB\left\{
 \sup_{m\ge1}\norm{\what\Phi_{m,n}(\btheta_m)-\Phi_m(\btheta_m)}_\infty
 \le K_\gM[\omega_R(h_{n,\delta})+h_{n,\delta}]
 \right\}\ge1-\delta,
\end{equation}
where $h_{n,\delta}$ is defined in Theorem~\ref{thm:A4-Winf}.
\end{lemma}

As in Section~\ref{subsec:proof-A3}, the contraction of $\what\Phi_{m,n}$ gives
\begin{equation*}
 \sup_{m\ge1}\Wbar_\infty(\Beta_m^{(n)},\Beta_m)
 \le\frac1{1-\gamma}
 \sup_{m\ge1}\norm{\what\Phi_{m,n}(\btheta_m)-\Phi_m(\btheta_m)}_\infty.
\end{equation*}
Therefore Lemma~\ref{lem:A4-Winf-one-step} proves Theorem~\ref{thm:A4-Winf}. 

\subsection{Analysis of the Local Asymptotic Minimax Risk}\label{subsec:proof-local}
To prove Theorem~\ref{thm:local-exact}, we establish matching upper and lower bounds for the local minimax risk. For the upper bound, we show that the expected Wasserstein loss of the quantile fixed point estimator, multiplied by $\sqrt n$, converges to the Gaussian risk uniformly over each local Hellinger neighborhood. For the lower bound, we construct a parametric submodel and derive the matching local asymptotic minimax bound from its Gaussian limit experiment.

First, we show that any estimator can be replaced by one with $m$ equally weighted atoms at each state, without increasing its Wasserstein loss to $\Beta_m(Q)$.
\begin{lemma}\label{lem:local-decision}
For $\mu\in\Delta(I)$, define
\[
 a_i(\mu)=m\int_{(i-1)/m}^{i/m}F_\mu^{-1}(u)\,\rd u,
 \qquad \gA_m\mu=\frac1m\sum_i\delta_{a_i(\mu)}.
\]
The $a_i(\mu)$ are ordered, and, for every ordered $\theta_1,\ldots,\theta_m$ and $1\le p\le\infty$,
\[
 W_p\left(\gA_m\mu,\frac1m\sum_i\delta_{\theta_i}\right)
 \le W_p\left(\mu,\frac1m\sum_i\delta_{\theta_i}\right).
\]
\end{lemma}
The proof is in Appendix~\ref{app:proof-local-decision}. Applying this map at each state reduces the loss of any distribution estimator, so it suffices to establish the lower bound for estimators in $\sP_m^{\gS}$.

For the upper bound, we need Gaussian convergence of the quantile estimation error and uniform integrability of its norm. The following lemma supplies both results and its proof is in Appendix~\ref{app:proof-joint-local-linearization}.
\begin{lemma}\label{lem:joint-local-linearization}
Fix $0<c<\infty$. For $Q\in\gN_n(c;Q_0)$, write
\[
 \bU_{n,Q}=\sqrt n\{\btheta_m^{(n)}-\btheta_m(Q)\}.
\]
Then, as $n\to\infty$,
\[
 \sup_{Q\in\gN_n(c;Q_0)}d_{\mathrm{BL}}\bigl(\mathcal L_Q(\bU_{n,Q}),\normal(0,\bV_m)\bigr)\to0,
\]
where the bounded Lipschitz distance is defined by
\[
 d_{\mathrm{BL}}(\mu,\nu)=\sup_{\substack{\|h\|_\infty\le1\\ \operatorname{Lip}(h)\le1}}\left|\int h\,\rd\mu-\int h\,\rd\nu\right|,
\]
with the Lipschitz constant taken in the Euclidean norm. Moreover, for every $1\le p\le\infty$, the family
\[
 \left\{\ell_{m,p}(\bU_{n,Q}):n\ge1,\ Q\in\gN_n(c;Q_0)\right\}
\]
is uniformly integrable under the respective laws $\PB_Q^{(n)}$.
\end{lemma}

To prove the upper bound, we apply the bounded Lipschitz convergence to bounded truncations of $\ell_{m,p}$. Uniform integrability allows the truncation level to tend to infinity, giving convergence of the expected norm uniformly to $\mathfrak C_{m,p}(Q_0)$. By Equation~\eqref{eq:Wp-array}, this proves Equation~\eqref{eq:local-attainment}, therefore both limiting minimax risks are at most $\mathfrak C_{m,p}(Q_0)$.

To construct the submodel for the lower bound, write $b=(s,a)$, $Q_b=Q_{s,a}$, and $X_{k,b}=X_k^{(s,a)}$. Using $y_{m,s,i}$ from Section~\ref{subsec:local-model}, set
\[
 [\bY_{m,(s,a)}(r,s^\prime)]_{u,i}
 =\ind\{u=s\}\pi(a\mid s)y_{m,s,i}(r,s^\prime),
 \qquad
 \bxi_{m,b}=\bY_{m,b}-\E_{Q_{0,b}}\bY_{m,b}.
\]
Define
\begin{equation}\label{eq:local-canonical}
 \bphi_{m,b}=-\bG_m^{-1}\bxi_{m,b},
 \qquad
 \bV_m=\sum_b\E_{Q_{0,b}}[\bphi_{m,b}\bphi_{m,b}^{\top}].
\end{equation}
The functions $\bphi_{m,b}$ determine perturbations of the sampling distributions that retain the full covariance $\bV_m$. The following lemma describes the resulting change in the target and places the local alternatives in the Hellinger neighborhoods. Its proof is in Appendix~\ref{app:proof-local-LF}.
\begin{lemma}\label{lem:local-LF}
Let $r=\operatorname{rank}(\bV_m)>0$. There exist an $r$-dimensional parametric submodel $\{Q_{\bt}\}\subset\gQ$, defined near $\bt=0$ with $Q_{\mathbf0}=Q_0$, and a matrix $\bA\in\RB^{d\times r}$ with $\bA\bA^\top=\bV_m$ such that, for every fixed $\bt\in\RB^r$,
\[
 \sqrt n\brc{\btheta_m\prn{Q_{\bt/\sqrt n}}-\btheta_m}\to\bA\bt.
\]
Moreover, the submodel is locally asymptotically normal at $\bt=0$, with Fisher information matrix $\bI_r$.
For every finite set $T\subset\RB^r$, there is $c_T<\infty$ such that $Q_{\bt/\sqrt n}\in\gN_n(c_T;Q_0)$ for all $\bt\in T$ and all sufficiently large $n$.
\end{lemma}

By Lemma~\ref{lem:local-decision} and~\ref{lem:local-LF}, 
\[
r_{n,m,p}(c_T;Q_0)\ge\inf_{\what{\btheta}_n}\max_{\bt\in T}\E_{Q_{\bt/\sqrt n}}\ell_{m,p}\prn{\sqrt n\brc{\what{\btheta}_n-\btheta_m(Q_{\bt/\sqrt n})}},
\]
where the infimum is over all $\RB^d$-valued estimators.

For $K>0$, the truncated loss $L_K(\bz)=\min\{\ell_{m,p}(\bz),K\}$ is bounded, symmetric, and has convex sublevel sets. The local asymptotic minimax theorem \citep[Theorem~8.11]{van2000asymptotic} therefore yields
\[
\lim_{c\to\infty}\liminf_{n\to\infty}r_{n,m,p}(c;Q_0)\ge \E L_K(\bA \bZ),
\qquad \bZ\sim\normal(0,\bI_r).
\]
Since $\bA\bA^\top=\bV_m$, letting $K\to\infty$ gives
\[
\underline{\gR}^{\mathrm{loc}}_{m,p}(Q_0)\ge\E\ell_{m,p}(\bA \bZ)=\mathfrak C_{m,p}(Q_0).
\]
Together with the upper bound, this proves \eqref{eq:local-exact}. If $\bV_m=0$, the conclusion follows directly from the upper bound and nonnegativity of the risk.

\subsection{Analysis of Sharpness of the Wasserstein Rates}\label{subsec:proof-instances}
To establish a lower bound for the Gaussian risk, we use the connection between the covariance of the canonical gradient and the sensitivity of the quantile fixed point to changes in the reward distributions. In particular, if a perturbation produces a non-vanishing change in the averaged quantile parameters at some state, then the asymptotic variances must be bounded away from zero. We construct such a perturbation uniformly over $m$, leading to the following lemma, the proof of which is in Appendix~\ref{app:proof-complexity-positive}.
\begin{lemma}\label{lem:complexity-positive}
There exists $c_{Q_0}>0$ such that
\[
 \inf_{m\ge1}\mathfrak C_{m,1}(Q_0)\ge c_{Q_0}.
\]
\end{lemma}

To prove Proposition~\ref{prop:uniform-instance}, it remains to verify Assumption~\ref{ass:A2}. The choice $\gamma=e^{-1}$ makes this an algebraic question. We use the following lemma, proved in Appendix~\ref{app:proof-uniform-boundary}.
\begin{lemma}\label{lem:uniform-boundary}
Consider an MDP with one state, one action, uniform rewards on $[0,1]$, and a transcendental discount $\gamma\in(0,1)$. For every $m\ge1$ and $i,j\in[m]$, we have
\[
 \theta_m(i)-\gamma\theta_m(j)\notin\{0,1\}.
\]
\end{lemma}

Thus the instance in Proposition~\ref{prop:uniform-instance} satisfies Assumptions~\ref{ass:A1}--\ref{ass:A3}. Theorem~\ref{thm:local-exact} and the expectation bound in Theorem~\ref{thm:A3} give, for each $m$,
\[
 \mathfrak C_{m,\infty}(Q_0)
 =\lim_{n\to\infty}\sqrt n\,\E_{Q_0}
   \Wbar_\infty(\Beta_m^{(n)},\Beta_m)\le C_{Q_0}.
\]
Together with Lemma~\ref{lem:complexity-positive} and
$\ell_{m,1}\le\ell_{m,p}\le\ell_{m,\infty}$, this proves
$\mathfrak C_{m,p}\asymp1$ uniformly in $m,p$.

To prove Theorem~\ref{thm:complexity-hard}, we use an MDP with two states $s,t$, one action, $\gamma=1/2$, and $P(t\mid s)=P(t\mid t)=1$. The return at $s$ consists of its immediate reward followed by a discounted return from $t$. We choose the reward density at $s$ to be extremely small on a sequence of intervals near zero, while a density comparable to that of a Beta$(2,2)$ distribution at $t$ gives control of the probabilities of small continuation returns. Set
\[
 x_j=8^{-j},\qquad \epsilon_j=\exp(-e^{j^2}).
\]
The following lemma constructs the reward densities; its proof is in Appendix~\ref{app:proof-hard-rewards}.

\begin{lemma}\label{lem:hard-rewards}
There exist reward densities $p_s,p_t$ and constants $c,C>0$ such that the resulting MDP $Q_*$ satisfies Assumptions~\ref{ass:A1},~\ref{ass:A2}, and~\ref{ass:A4}, and for all sufficiently large $j$,
\begin{align*}
 c\epsilon_j\le p_s(r)\le C\epsilon_j,&\qquad x_j/4\le r\le x_j,\\
 p_s(r)\le C\epsilon_j,&\qquad 0\le r\le x_j,\\
 cr(1-r)\le p_t(r)\le Cr(1-r),&\qquad 0\le r\le1.
\end{align*}
\end{lemma}

Fix this MDP for the rest of the proof. All population quantities below refer to $Q_*$, and the positive constants $c,C,K$ may change from line to line but depend only on $Q_*$. Let $\nu=\eta^\pi(t)$ and $\nu_m=\eta_m(t)$, and define
\[
 H(x)=\PB(R_s+\gamma Z\le x),\qquad
 H_m(x)=\PB(R_s+\gamma Z_m\le x),
 \qquad Z\sim\nu,\quad Z_m\sim\nu_m,
\]
where $R_s$ has density $p_s$ and is independent of $Z$ and $Z_m$. The deterministic transitions give $H_m=F_{m,s}$.

We choose the resolutions to place the smallest quantile at $s$ inside the intervals where $p_s$ is small. Set $a_j=H(3x_j/4)$ and let $m_j$ be the integer nearest to $(2a_j)^{-1}$. Then the first quantile level $1/(2m_j)$ is close to $a_j$. To locate the corresponding quantile of $H_{m_j}$, we use the bound $W_1(\nu_m,\nu)\le C/m$. Since $p_s\le C\epsilon_j$ on $[0,x_j]$, replacing $\nu$ by $\nu_m$ changes $H$ by at most $C\epsilon_j/m$ on this interval. The probability of small returns at $t$ provides the increase in $H$ needed to keep the quantile between $x_j/2$ and $x_j$. These estimates yield the following lemma, proved in Appendix~\ref{app:proof-hard-resolutions}.

\begin{lemma}\label{lem:hard-resolutions}
For the resolutions chosen above, let $q_j=\theta_{m_j}(s,1)=H_{m_j}^{-1}(1/(2m_j))$. For all sufficiently large $j$, then
\[
 q_j\in[x_j/2,x_j],\qquad
 \nu_{m_j}([0,x_j/4])\ge e^{-Cj^2},
\]
and
\[
 \frac{c}{\epsilon_jx_j}\le m_j
 \le\frac{Ce^{Cj^2}}{\epsilon_jx_j},
 \qquad \log m_j=e^{j^2}+O(j^2).
\]
\end{lemma}

We next compare the sampling variability at $q_j$ with the target density there. For a reward value $r$, write
\[
 g_j(r)=F_{\nu_{m_j}}\!\left(\frac{q_j-r}{\gamma}\right).
\]
This is the probability that $r+\gamma Z_{m_j}$ does not exceed $q_j$. Thus, the CDF of $[\widehat{\gT}_n^\pi\Beta_{m_j}](s)$ at $q_j$ is $n^{-1}\sum_{k=1}^n g_j(R_k^{(s)})$. On $r\in[x_j/4,3x_j/8]$, the value of $g_j(r)$ is at least $\nu_{m_j}([0,x_j/4])$. The reward density bounds and Lemma~\ref{lem:hard-resolutions} therefore give the following estimates, proved in Appendix~\ref{app:proof-hard-cdf-variance}.

\begin{lemma}\label{lem:hard-cdf-variance}
For $R_s$ with density $p_s$ and all sufficiently large $j$, we have that
\[
 \var(g_j(R_s))\ge c\epsilon_jx_j e^{-Cj^2},
 \qquad f_{m_j,s}(q_j)\le C\epsilon_j.
\]
\end{lemma}

Because both states transition to $t$, the matrix $\bG_m^{-1}$ is block upper triangular, with a diagonal block corresponding to state $s$. Hence, omitting the single action from the block index, the contribution of samples at $s$ to the influence function in \eqref{eq:local-canonical} is
\[
 [\bphi_{m_j,s}(R_s,t)]_{s,1}
 =-\frac{g_j(R_s)-1/(2m_j)}{f_{m_j,s}(q_j)}.
\]
The samples at the two states are independent, so their contributions to the variance add. Lemma~\ref{lem:hard-cdf-variance} gives
\[
 [\bV_{m_j}]_{(s,1),(s,1)}
 \ge\frac{\var(g_j(R_s))}{f_{m_j,s}(q_j)^2}
 \ge c\frac{x_j}{\epsilon_j}e^{-Cj^2}.
\]
By the upper bound on $m_j$ in Lemma~\ref{lem:hard-resolutions}, the last expression is at least $c m_j x_j^2e^{-C'j^2}$. Because $x_j=8^{-j}$ and $\log m_j=e^{j^2}+O(j^2)$, this yields
\begin{equation*}
 [\bV_{m_j}]_{(s,1),(s,1)}
 \ge\frac{c m_j}{(\log m_j)^K}.
\end{equation*}
Let $\bZ_m\sim\normal(0,\bV_m)$. The contribution of the coordinate $(s,1)$ then gives, for $2<p\le\infty$,
\begin{align*}
 \mathfrak C_{m_j,p}(Q_*)
 &\ge m_j^{-1/p}\E|Z_{m_j}(s,1)|\\
 &=\sqrt{\frac2\pi}\,m_j^{-1/p}
       \sqrt{[\bV_{m_j}]_{(s,1),(s,1)}}
 \ge\frac{c\,m_j^{1/2-1/p}}{(\log m_j)^{K/2}}.
\end{align*}
For the upper bound, integrating \eqref{eq:Vn-exp-tail} bounds the second moment uniformly in $m$. Passing to the Gaussian limit at each fixed $m$, using bounded truncations, yields
\[
 \E\ell_{m,2}(\bZ_m)^2
 \le\liminf_{n\to\infty}n\E_{Q_*}
      \Wbar_2(\Beta_m^{(n)},\Beta_m)^2
 \le C(Q_*)^2.
\]
The norm comparison used to prove Corollary~\ref{cor:A4-rootn} then gives
\[
 \mathfrak C_{m,p}(Q_*)
 \le m^{1/2-1/p}\{\E\ell_{m,2}(\bZ_m)^2\}^{1/2}
 \le C(Q_*)m^{1/2-1/p}.
\]
For $1\le p\le2$, use $\ell_{m,p}\le\ell_{m,2}$ and Lemma~\ref{lem:complexity-positive}. This completes the proof of Theorem~\ref{thm:complexity-hard}.

%% file: sections/conclusion.tex
In this paper, we have studied the statistical accuracy of the quantile fixed point estimator under Wasserstein distances. Although a finer quantile representation brings more parameters and increasingly extreme quantile levels, it does not certainly lead to greater estimation error. Under uniformly positive reward densities, we have established the estimation error bounds of order $n^{-1/2}$ uniformly over all resolutions $m$ in the $p$-th Wasserstein distance for each $p\ge 1$. Even when the reward densities vanish at the support endpoints, this resolution-independent rate persists in the \(\Wbar_p\) distances for $1\leq p\leq2$. For $p>2$, we have established both $O(n^{-1/2})$ bounds with polynomial dependence on $m$ and slower bounds that remain uniform at all resolutions.

To assess the sharpness of these bounds, we have characterized the exact local asymptotic minimax risk at each fixed resolution and shown that it is attained by the quantile fixed point estimator. We have also used two examples to examine how this optimal risk depends on the resolution. In the first example, with a uniformly positive reward density, the optimal local risk is bounded above and away from zero uniformly over all resolutions and Wasserstein orders. In the second, with endpoint-vanishing reward densities, the optimal local risk can grow at the polynomial order predicted by our non-asymptotic bounds for $p>2$, up to logarithmic factors along a sequence of resolutions. Together, these results show that the statistical cost of increasing quantile resolution depends fundamentally on the reward density and the Wasserstein order.

%% file: appendices/key_lemmas.tex
In this section, we prove the lemmas stated in Section~\ref{sec:strategy}. 

\subsection{Proof of Lemma~\ref{lem:A3-one-step}}\label{app:proof-A3-group}
\begin{proof}
For each $(s,a,s^\prime)$, define
\[
N_{s,a,s^\prime}
=\sum_{k=1}^n
\ind\{S_k^{\prime(s,a)}=s^\prime\},
\qquad
\what P(s^\prime\mid s,a)
=\frac{N_{s,a,s^\prime}}{n}.
\]
For
$N_{s,a,s^\prime}>0$, let
\[
 \what F_{s,a,s^\prime}(x)
 =\frac1{N_{s,a,s^\prime}}\sum_{k=1}^n
 \ind\{S_k^{\prime(s,a)}=s^\prime,\ R_k^{(s,a)}\le x\}.
\]
With $\gI=\{(s,a,s^\prime):P(s^\prime\mid s,a)>0\}$, define
\[
 d_n=\max_{(s,a,s^\prime)\in\gI}
 \norm{\what F_{s,a,s^\prime}-F_{s,a}}_\infty,
 \qquad
 r_n=\max_{(s,a,s^\prime)\in\gI}
 \abs{\frac{\what P(s^\prime\mid s,a)}{P(s^\prime\mid s,a)}-1},
\]
where $\what F_{s,a,s^\prime}$ is the CDF of a point mass at zero when the stratum
is empty, as specified before Lemma~\ref{lem:primitive-conc}. 

Define the hybrid joint law by
\[
\wtilde Q_{s,a}^{(n)}(\rd(r,s^\prime))=\what P(s^\prime\mid s,a)\times\gP_R(\rd r\mid s,a),
\]
and the corresponding Bellman operator by
\[
[\wtilde{\gT}_n^\pi\bm\eta](s)
=\sum_{a\in\gA}\pi(a\mid s)
\int (b_{r,\gamma})_\#\eta(s^\prime)\,
\wtilde Q_{s,a}^{(n)}(\rd(r,s^\prime)).
\]
Write $\wtilde F_{m,s}$ for the CDF of $\wtilde{\gT}_n^\pi\bm\eta_m$ at state $s$, and define the three
Bellman-target CDFs
\begin{align*}
 \what F_{s,\btheta_m}^{(n)}(x)
 &=\sum_{a,s^\prime,j}\frac{\pi(a\mid s)\what P(s^\prime\mid s,a)}m
      \what F_{s,a,s^\prime}(x-\gamma\theta_m(s^\prime,j)),\\
 \wtilde F_{m,s}(x)
 &=\sum_{a,s^\prime,j}\frac{\pi(a\mid s)\what P(s^\prime\mid s,a)}m
      F_{s,a}(x-\gamma\theta_m(s^\prime,j)),\\
 F_{m,s}(x)
 &=\sum_{a,s^\prime,j}\frac{\pi(a\mid s)P(s^\prime\mid s,a)}m
      F_{s,a}(x-\gamma\theta_m(s^\prime,j)).
\end{align*}
For every nonempty stratum, the definition of $d_n$ and generalized-inverse bracketing give
\[
 F_{s,a}^{-1}((u-d_n)\vee0)
 \le \what F_{s,a,s^\prime}^{-1}(u)
 \le F_{s,a}^{-1}((u+d_n)\wedge1),\qquad 0<u<1.
\]
Under Assumption~\ref{ass:A3}, $p_{s,a}\ge c_0$, so $F_{s,a}^{-1}$ is $1/c_0$-Lipschitz, including at the endpoints. Consequently,
\[
 \sup_{0<u<1}\abs{\what F_{s,a,s^\prime}^{-1}(u)
                    -F_{s,a}^{-1}(u)}\le\frac{d_n}{c_0}.
\]
To couple the first two targets, sample $A\sim\pi(\cdot\mid s)$,
$S^\prime\mid A\sim\what P(\cdot\mid s,A)$,
$J\sim\mathrm{Unif}\{1,\ldots,m\}$, and $U\sim\mathrm{Unif}(0,1)$,
with $J,U$ independent of $(A,S^\prime)$. Set
\[
 \what X=\what F_{s,A,S^\prime}^{-1}(U)+\gamma\theta_m(S^\prime,J),
 \qquad
 \wtilde X=F_{s,A}^{-1}(U)+\gamma\theta_m(S^\prime,J).
\]
The variables $\what X$ and $\wtilde X$ have the first and second CDFs
above, respectively. Therefore,
\[
 \abs{\what X-\wtilde X}
 =\abs{\what F_{s,A,S^\prime}^{-1}(U)-F_{s,A}^{-1}(U)}
 \le\frac{d_n}{c_0}\quad\text{a.s.}
\]
Taking the supremum over states proves
\begin{equation}\label{eq:A3-reward-pert}
 \Wbar_\infty(\what{\gT}_n^\pi\Beta_m,\wtilde{\gT}_n^\pi\Beta_m)
 \le \frac{d_n}{c_0}.
\end{equation}

To compare the second and third CDFs, set
\[
 w_{a,s^\prime,j}=\frac{\pi(a\mid s)P(s^\prime\mid s,a)}m,
 \qquad
 \what w_{a,s^\prime,j}=\frac{\pi(a\mid s)\what P(s^\prime\mid s,a)}m.
\]
By the definition of $r_n$,
\[
 \abs{\what w_{a,s^\prime,j}-w_{a,s^\prime,j}}
 \le r_n w_{a,s^\prime,j},
 \qquad
 \sum_{a,s^\prime,j}(\what w_{a,s^\prime,j}-w_{a,s^\prime,j})=0.
\]
Consequently, writing $F=F_{m,s}$ and $\wtilde F=\wtilde F_{m,s}$,
\begin{align*}
 \abs{\wtilde F(x)-F(x)}
 &\le \sum_{a,s^\prime,j}\abs{\what w_{a,s^\prime,j}-w_{a,s^\prime,j}}
          F_{s,a}(x-\gamma\theta_m(s^\prime,j))
 \le r_n F(x),\\
 \abs{\wtilde F(x)-F(x)}
 &=\abs{\sum_{a,s^\prime,j}(\what w_{a,s^\prime,j}-w_{a,s^\prime,j})
          [1-F_{s,a}(x-\gamma\theta_m(s^\prime,j))]}
 \le r_n[1-F(x)].
\end{align*}
Thus
\begin{equation}\label{eq:transition-relative-CDF}
 \abs{\wtilde F(x)-F(x)}\le r_n\min\{F(x),1-F(x)\}.
\end{equation}
Suppose first that $0<r_n\le1/8$.
Fix $u=\tau_i$ and put $q=F^{-1}(u)$. Set
\[
 d_u=4r_nu(1-u),\qquad u_-=u-d_u,\qquad u_+=u+d_u,
\]
then $u_-,u_+\in(0,1)$. The function $a(v)=\min\{v,1-v\}$ is $1$-Lipschitz and satisfies $a(u)\le2u(1-u)$. Hence $(1-r_n)d_u>r_na(u)$, and therefore
\[
 u_-+r_na(u_-)<u,\qquad u_+-r_na(u_+)>u.
\]
Continuity gives $F(F^{-1}(u_\pm))=u_\pm$. Substituting these points into \eqref{eq:transition-relative-CDF} yields
\[
 \wtilde F(F^{-1}(u_-))<u<\wtilde F(F^{-1}(u_+)),
 \qquad
 F^{-1}(u_-)\le\wtilde F^{-1}(u)\le F^{-1}(u_+).
\]
By Lemma B.1 in~\citet{cheng2026statistical}, there is a constant $c_\gM>0$ only depending on $\gM$, such that
\begin{equation*}
 |F(q+h)-u|\ge c_\gM u(1-u)|h|,\qquad 0<|h|\le B.
\end{equation*}
Applying this inequality at $h=F^{-1}(u_\pm)-q$ gives
\[
 |F^{-1}(u_\pm)-q|\le
 \frac{|u_\pm-u|}{c_\gM u(1-u)}=\frac{4r_n}{c_\gM}.
\]
Consequently the same bound holds for $|\wtilde F^{-1}(\tau_i)-F^{-1}(\tau_i)|$. Taking the maximum over states and coordinates proves
\begin{equation}\label{eq:A3-transition-pert}
 \Wbar_\infty(\bPi_m\wtilde{\gT}_n^\pi\Beta_m,
                  \bPi_m\gT^\pi\Beta_m)\le C_\gM r_n.
\end{equation}
For $r_n>1/8$, use the support bound $B\le8Br_n$ instead, and for $r_n=0$ we have $F_{m,s}=\wtilde F_{m,s}$. Thus Equation~\eqref{eq:A3-transition-pert} always holds with an instance-dependent constant for all $m$.

$\bPi_m$ is nonexpansive in $\Wbar_\infty$, so \eqref{eq:A3-reward-pert} and \eqref{eq:A3-transition-pert} imply that
\[
 \Delta_n\le C_\gM(d_n+r_n).
\]
Finally, define
\[
 L_{\gI}\coloneq \max\left\{1,\frac{\log(2|\gI|)}{\log(4|\gS|)}\right\}.
\]
For every $\delta\in(0,1)$,
\[
 \log\frac{2|\gI|}{\delta}
 \le L_{\gI}\log\frac{4|\gS|}{\delta}.
\]
Combining this inequality with Lemma~\ref{lem:primitive-conc} yields
\[
 \PB\left(
 \Delta_n>K_\gM\sqrt{\frac{\log(4|\gS|/\delta)}{n}}
 \right)\le\delta,
\]
with
\[
 K_\gM=\frac{5C_\gM}{p_*}\sqrt{\frac{L_{\gI}}2}.
\]
Its expectation estimate also gives
\[
 \E\Delta_n\le\frac{5C_\gM}{p_*}\sqrt{\frac{1+\log(2|\gI|)}{2n}},
\]
which proves the expectation bound.
\end{proof}

\subsection{Proofs of Lemmas~\ref{lem:common-localization}--\ref{lem:joint-bernstein-event}}\label{app:proof-A4-W2-group}
\phantomsection\label{app:proof-common-localization}
\begin{proof}[Proof of Lemma~\ref{lem:common-localization}]
We first establish the boundary monotonicity and interior density lower bound. For $0\le y\le x\le\ell$ and every $\theta(s^\prime,j)\ge0$, both shifted reward arguments $x-\gamma\theta(s^\prime,j)$ and $y-\gamma\theta(s^\prime,j)$ lie in $(-\infty,\ell]$, where the zero-extended reward density is nondecreasing. Similarly, for $B-\ell\le x\le y\le B$ and $\theta(s^\prime,j)\le B$, the arguments lie in $[1-\ell,\infty)$, where it is nonincreasing. Summing the nonnegative weights in \eqref{eq:target-density} gives
\begin{equation}\label{eq:boundary-mono}
 f_{s,\btheta}(y)\le f_{s,\btheta}(x)\quad(0\le y\le x\le\ell),
 \qquad
 f_{s,\btheta}(y)\le f_{s,\btheta}(x)\quad(B-\ell\le x\le y\le B).
\end{equation}
By Proposition~\ref{prop:return_density},
\[
 c_{\mathrm{ret}}(\gM,\ell)\coloneq
 \min_s\inf_{x\in[\ell,B-\ell]}p_{\eta^\pi(s)}(x)>0.
\]
For each successor state $s^\prime$, couple $Y\sim\eta_{\btheta}(s^\prime)$ and $Y^\prime\sim\eta^\pi(s^\prime)$ optimally in $W_1$. For $d>0$,
\begin{align*}
 \abs{\E p_{s,a}(x-\gamma Y)-\E p_{s,a}(x-\gamma Y^\prime)}
 &\le\varpi_R(d)+2C_0\PB(\gamma|Y-Y^\prime|>d)\\
 &\le\varpi_R(d)+\frac{2C_0\gamma}{d}
 W_1(\eta_{\btheta}(s^\prime),\eta^\pi(s^\prime)).
\end{align*}
Averaging over $(a,s^\prime)$ with weights $\pi(a\mid s)P(s^\prime\mid s,a)$ yields, uniformly in $m,s,x$ and $\btheta\in\gU_m(\rho_0)$,
\[
 \abs{f_{s,\btheta}(x)-p_{\eta^\pi(s)}(x)}
 \le\varpi_R(d)+\frac{2C_0\gamma\rho_0}{d}.
\]
Choose $d>0$ so that $\varpi_R(d)\le c_{\mathrm{ret}}(\gM,\ell)/4$, and then choose $\rho_0>0$ so that $2C_0\gamma\rho_0/d\le c_{\mathrm{ret}}(\gM,\ell)/4$. This proves the interior lower bound with $c_{\gM,\ell}=c_{\mathrm{ret}}(\gM,\ell)/2$. Since $p_{s,a}\leq C_0$ and $p_{s,a}(0)=p_{s,a}(1)=0$, we have $f_{s,\btheta}\le C_0$ and $f_{s,\btheta}(0)=f_{s,\btheta}(B)=0$.

Moreover, for any $\rho>0$, the neighborhood $\gU_m(\rho)$ is convex. Indeed, if $\btheta,\bvartheta\in\gU_m(\rho)$ and $\btheta_t=(1-t)\btheta+t\bvartheta$, then coordinatewise interpolation also interpolates the quantile functions on each probability cell, and
\begin{align*}
 W_1(\eta_{\btheta_t}(s),\eta^\pi(s))
 &=\int_0^1\abs{(1-t)F_{\eta_{\btheta}(s)}^{-1}(u)
     +tF_{\eta_{\bvartheta}(s)}^{-1}(u)-F_{\eta^\pi(s)}^{-1}(u)}\,\rd u\\
 &\le(1-t)W_1(\eta_{\btheta}(s),\eta^\pi(s))
       +tW_1(\eta_{\bvartheta}(s),\eta^\pi(s))\le\rho.
\end{align*}
Taking the maximum over states proves the claim.

We next prove localization. For any probability law $\mu$ on $I$, write $q=F_\mu^{-1}$. Monotonicity gives
\[
 W_1(\Pi_m\mu,\mu)
 =\sum_{i=1}^m\int_{(i-1)/m}^{i/m}|q(u)-q(\tau_i)|\,\rd u
 \le\frac1{2m}\sum_{i=1}^m
       \{q(i/m)-q((i-1)/m)\}
 \le\frac{B}{2m}.
\]
Since $\Beta_m=\bPi_m\gT^\pi\Beta_m$, the $\Wbar_1$ contraction yields
\[
 \Wbar_1(\Beta_m,\Beta^\pi)
 \le\frac{B}{2m}+\gamma\Wbar_1(\Beta_m,\Beta^\pi),
\]
and hence
\begin{equation}\label{eq:basic-approx}
 \Wbar_1(\Beta_m,\Beta^\pi)\le\frac{B}{2m(1-\gamma)}.
\end{equation}
By the definitions of the Bellman CDFs,
\begin{equation}\label{eq:joint-cdf-decomp}
 \what F_{s,\btheta}^{(n)}(x)-F_{s,\btheta}(x)
 =\sum_{a,s^\prime,j}\frac{\pi(a\mid s)}m
 (\what J_{s,a,s^\prime}-J_{s,a,s^\prime})
 (x-\gamma\theta(s^\prime,j)).
\end{equation}
Consequently, uniformly in $m$, $\btheta$ and $s$,
\[
 \norm{\what F_{s,\btheta}^{(n)}-F_{s,\btheta}}_\infty
 \le\sum_a\pi(a\mid s)\sum_{s^\prime}
 \norm{\what J_{s,a,s^\prime}-J_{s,a,s^\prime}}_\infty
 \le D_n.
\]
In particular, this holds at $\btheta=\btheta_m^{(n)}$.
Since $\btheta_m^{(n)}
=\what\Phi_{m,n}(\btheta_m^{(n)})$, we have
\[
 \what F_{s,\btheta_m^{(n)}}^{(n)}
   (\theta_m^{(n)}(s,i)-)
 \le\tau_i\le
 \what F_{s,\btheta_m^{(n)}}^{(n)}
   (\theta_m^{(n)}(s,i)).
\]
Taking the left limit in the CDF bound and using the
continuity of $F_{s,\btheta_m^{(n)}}$, we obtain
\[
 \left|
 F_{s,\btheta_m^{(n)}}(\theta_m^{(n)}(s,i))
 -\tau_i
 \right|\le D_n.
\]
For any $x\in[0,B]$, let
$k=\#\{i:\theta_m^{(n)}(s,i)\le x\}$.
When $1\le k<m$, monotonicity yields
\[
 \frac{k-1/2}{m}-D_n
 \le F_{s,\btheta_m^{(n)}}(x)
 \le\frac{k+1/2}{m}+D_n.
\]
For $k=0,m$, the same inequality follows from the bounds at the first and last quantiles, respectively.
Since $F_{\eta_m^{(n)}(s)}(x)=k/m$, we conclude that
\[
 \left\|
 F_{\eta_m^{(n)}(s)}
 -F_{s,\btheta_m^{(n)}}
 \right\|_\infty
 \le D_n+\frac1{2m}.
\]
Integrating over $[0,B]$ gives
\[
 \Wbar_1(\Beta_m^{(n)},\gT^\pi\Beta_m^{(n)})
 \le B\left(D_n+\frac1{2m}\right).
\]
Use $\Beta^\pi=\gT^\pi\Beta^\pi$ and the $\Wbar_1$ contraction to obtain
\[
 \Wbar_1(\Beta_m^{(n)},\Beta^\pi)
 \le\frac{B}{1-\gamma}\left(D_n+\frac1{2m}\right).
\]
For each $0<\rho\le\rho_0$, choose $M_1$ so that $B/[2M_1(1-\gamma)]\le\rho/2$, and choose $\varepsilon_0$ so that $B\varepsilon_0/(1-\gamma)\le\rho/2$. These choices prove the simultaneous localization.

Finally, fix $M\ge1$. By Lemma~A.2 in \citet{cheng2026statistical},
\[
 \min_{\substack{1\le m\le M\\s\in\gS,\ i\in[m]}}
 f_{s,\btheta_m}(\theta_m(s,i))>0.
\]
Assumptions~\ref{ass:A1}--\ref{ass:A2} also imply continuity of the target densities near these quantiles. Since there are finitely many coordinates, there exist $c_{M,\gM},h_{M,\gM}>0$ such that
\[
 f_{s,\btheta_m}(x)\ge c_{M,\gM}
 \quad\text{for }1\le m\le M,\ s\in\gS,\ i\in[m],\
 |x-\theta_m(s,i)|\le h_{M,\gM}.
\]
Set $d_{M,\gM}=c_{M,\gM}h_{M,\gM}/2$.
When $D_n\le d_{M,\gM}$, the inverse bracketing above and \eqref{eq:joint-cdf-decomp} give
\[
 \|\what\Phi_{m,n}(\btheta_m)-\Phi_m(\btheta_m)\|_\infty
 \le D_n/c_{M,\gM}.
\]
The empirical contraction then yields
\[
 \max_{1\le m\le M}\Wbar_\infty(\Beta_m^{(n)},\Beta_m)
 \le\frac{D_n}{(1-\gamma)c_{M,\gM}}.
\]
When $D_n>d_{M,\gM}$, use the support bound $\Wbar_\infty\le B\le (B/d_{M,\gM})D_n$. Increasing the constant covers both cases, and therefore Lemma~\ref{lem:common-localization} holds.
\end{proof}

\phantomsection\label{app:proof-W2-contraction}
\begin{proof}[Proof of Lemma~\ref{lem:W2-contraction}]
Choose $0<\ell<\min\{\kappa,1/2,B/2\}$ and $\zeta>0$ so that
\[
 L=1+4C_0\ell+\zeta,\qquad\gamma\sqrt L<1.
\]
Choose $0<\rho\le\rho_0$ from Lemma~\ref{lem:common-localization} and $M_0$ from Lemma~\ref{lem:column} for $\ell$ and $\zeta$. As shown in the proof of Lemma~\ref{lem:common-localization}, the neighborhood $\gU_m(\rho)$ is convex, so $\btheta_t=(1-t)\btheta+t\bvartheta$ remains in it. Put $q_{s,i}(t)=[\Phi_m(\btheta_t)]_{s,i}$ and write $\bq(t)=(q_{s,i}(t))_{s,i}$.

By Lemma~\ref{lem:common-localization} and \eqref{eq:boundary-mono},
$f_{s,\btheta_t}(q_{s,i}(t))>0$ for every $t$. The zero-extended reward densities are continuous by Assumptions~\ref{ass:A1} and~\ref{ass:A4}, hence their CDFs are $C^1$ on $\RB$, including at support endpoints. Thus $(t,x)\mapsto F_{s,\btheta_t}(x)$ is $C^1$. The implicit function theorem applied to
\[
 F_{s,\btheta_t}(q_{s,i}(t))=\tau_i
\]
shows that $q_{s,i}$ is $C^1$ on $(0,1)$. Differentiating the displayed equation gives
\[
  \dot q_{s,i}(t)
  =\gamma\sum_{a,s^\prime,j}
  \frac{\pi(a\mid s)P(s^\prime\mid s,a)}m
  \frac{p_{s,a}(q_{s,i}(t)-\gamma\theta_t(s^\prime,j))}
  {f_{s,\btheta_t}(q_{s,i}(t))}
  (\vartheta(s^\prime,j)-\theta(s^\prime,j)).
\]
The coefficient matrix after dividing by $\gamma$ is nonnegative and each row sums to one.  Hence Jensen's inequality gives
\begin{align*}
  \frac1m\sum_i|\dot q_{s,i}(t)|^2
  &\le\gamma^2
  \sum_{a,s^\prime,j}\frac{\pi(a\mid s)P(s^\prime\mid s,a)}m
  |\vartheta(s^\prime,j)-\theta(s^\prime,j)|^2\\
  &\qquad\times
  \left[\frac1m\sum_i
  \frac{p_{s,a}(q_{s,i}(t)-\gamma\theta_t(s^\prime,j))}
  {f_{s,\btheta_t}(q_{s,i}(t))}\right].
\end{align*}
By Lemma~\ref{lem:column}, for $m\geq M_0$,
\[
  \frac1m\sum_i
  \frac{p_{s,a}(q_{s,i}(t)-\gamma\theta_t(s^\prime,j))}
  {f_{s,\btheta_t}(q_{s,i}(t))}\leq L\coloneq 1+4C_0\ell+\zeta.
\]
Thus
\[
  \norm{\dot{\bq}(t)}_{2,m}
  \le\gamma\sqrt L\,\norm{\bvartheta-\btheta}_{2,m}.
\]
The choice at the start gives $\lambda_2=\gamma\sqrt L<1$. Therefore, 
\begin{align*}
 \norm{\Phi_m(\bvartheta)-\Phi_m(\btheta)}_{2,m}
 &\le\int_0^1\norm{\dot{\bq}(t)}_{2,m}\,\rd t\\
 &\le\lambda_2\norm{\bvartheta-\btheta}_{2,m}
\end{align*}
and Lemma~\ref{lem:W2-contraction} holds.
\end{proof}

\phantomsection\label{app:proof-square-inverse-additive}
\begin{proof}[Proof of Lemma~\ref{lem:square-inverse-additive}]
Fix $n,m,s$ and $\btheta\in\gU_m(\rho)$ as in the statement, and write
\[
 F=F_{s,\btheta},\qquad G=\what F_{s,\btheta}^{(n)},\qquad
 f=f_{s,\btheta},\qquad q=F^{-1},\qquad \what q=G^{-1}.
\]
Assumption~\ref{ass:A4}, Lemma~\ref{lem:common-localization} and \eqref{eq:boundary-mono} give the density bounds and boundary monotonicity needed below. If $\varepsilon\ge1/4$, the common support gives
\[
 \frac1m\sum_i|\what q(\tau_i)-q(\tau_i)|^2
 \le B^2\le4B^2\varepsilon.
\]
We henceforth suppose $0\le\varepsilon<1/4$. If $z=\varepsilon=0$, the two CDFs coincide and the assertion is immediate. 

For $\varepsilon<u<1-\varepsilon$, set $v_\pm=u\pm\varepsilon$ and $r(u)=q(v_+)-q(v_-)$.
By Lemma~\ref{lem:shifted-envelopes},
\[
 \abs{\what q(u)-q(u)}^2
 \le C_{\mathrm{geom}}\bigl[r(u)^2+z^2+H_L(v_-)+H_R(v_+)\bigr].
\]
Using \eqref{eq:envelope-integrals} and changing variables,
\begin{align*}
 &\int_\varepsilon^{1-\varepsilon}
   [H_L(u-\varepsilon)+H_R(u+\varepsilon)]\,\rd u\\
 =&\int_0^{1-2\varepsilon}H_L(v)\,\rd v
   +\int_{2\varepsilon}^1H_R(v)\,\rd v
 \le14\ell z^2.
\end{align*}
Moreover, since $r(u)\le B$, we have
\begin{align*}
 \int_\varepsilon^{1-\varepsilon}r(u)^2\,\rd u&\leq B\brk{\int_{2\varepsilon}^{1}q(v)\,\rd v-\int_0^{1-2\varepsilon}q(v)\,\rd v}\\
 &=B\brk{\int_{1-2\varepsilon}^{1}q(v)\,\rd v-\int_0^{2\varepsilon}q(v)\,\rd v}\le2B^2\varepsilon.
\end{align*}
Using $|\what q-q|\le B$ on the two remaining intervals of total length $2\varepsilon$, we obtain
\begin{equation}\label{eq:additive-continuum}
 W_2(G,F)^2\le C_{\mathrm{geom}}(z^2+\varepsilon),
\end{equation}
after increasing the constant depending only on $\gM$ and $\ell$. We next pass from this integral bound to the midpoint grid.

Suppose first that $m\varepsilon<1/4$. Set
$v_{i,\pm}=\tau_i\pm\varepsilon$. Then
\[
 \frac{i-1}{m}+\frac1{4m}
 <v_{i,-}\le v_{i,+}
 <\frac{i}{m}-\frac1{4m}.
\]
By Lemma~\ref{lem:shifted-grid} and
\eqref{eq:envelope-integrals},
\begin{align*}
 &\frac1m\sum_{i=1}^m
 [H_L(v_{i,-})+H_R(v_{i,+})]\\
 \le&4\int_0^1(H_L+H_R)(v)\,\rd v
 \le56\ell z^2.
\end{align*}

By \eqref{eq:rank-shift-average},
\[
 \frac1m\sum_{i=1}^m r(\tau_i)^2
 \le B\prn{16\ell+\frac 2{c_{\gM,\ell}}}\varepsilon.
\]
Combining these bounds with
\eqref{eq:shifted-pointwise} yields
\[
 \frac1m\sum_{i=1}^m\abs{\what q(\tau_i)-q(\tau_i)}^2\le C_{\mathrm{geom}}\left[(1+56\ell)z^2+B\prn{16\ell+\frac{2}{c_{\gM,\ell}}}\varepsilon\right].
\]

Now we suppose $m\varepsilon\ge1/4$. For any law $\mu$ supported in $[0,B]$, the midpoint quantization estimate in the proof of Lemma~\ref{lem:common-localization} and the support diameter imply
\[
 W_2(\Pi_m\mu,\mu)^2\le B W_1(\Pi_m\mu,\mu)\le\frac{B^2}{2m}.
\]
Apply the triangle inequality through $G$ and $F$ and use \eqref{eq:additive-continuum}:
\begin{align*}
 m^{-1}\sum_i\abs{G^{-1}(\tau_i)-F^{-1}(\tau_i)}^2&=W_2(\Pi_mG,\Pi_mF)^2\\
 &\le3W_2(G,F)^2+\frac{3B^2}{m}\\
 &\le C_{\mathrm{geom}}(z^2+\varepsilon),
\end{align*}
since $1/m\le4\varepsilon$. Taking $C_{\gM,\ell}$ to dominate the constants in these cases proves \eqref{eq:additive-inverse-result}.
\end{proof}

\phantomsection\label{app:proof-joint-bernstein-event}
\begin{proof}[Proof of Lemma~\ref{lem:joint-bernstein-event}]
We use the notations in Appendix~\ref{app:proof-A3-group}. 
First, set
\[
 Z_{s,a,s^\prime}=\norm{\what J_{s,a,s^\prime}-J_{s,a,s^\prime}}_\infty,
 \qquad D_{n,s}=\sum_a\pi(a\mid s)\sum_{s^\prime}Z_{s,a,s^\prime},
\]
so that $D_n=\max_sD_{n,s}$. For each triple, replace $R_k^{(s,a)}$ by $2$ whenever $S_k^{\prime(s,a)}\ne s^\prime$. The CDF of the resulting variable agrees with $J_{s,a,s^\prime}$ on $[0,1]$, so the DKW inequality gives
\[
 \PB(Z_{s,a,s^\prime}>v)\le2e^{-2nv^2},\qquad v>0.
\]
In particular,
\begin{align*}
 \E\exp\!\left(\frac n2Z_{s,a,s^\prime}^2\right)
 &=1+\frac n2\int_0^\infty e^{nv/2}\PB(Z_{s,a,s^\prime}^2>v)\,\rd v\\
 &\le1+n\int_0^\infty e^{-3nv/2}\,\rd v=\frac53.
\end{align*}
For fixed $s$, the weights $\lambda_{a,s^\prime}=\pi(a\mid s)/|\gS|$ sum to one. Jensen's inequality gives
\begin{align*}
 \E\exp\!\left(\frac{nD_{n,s}^2}{2|\gS|^2}\right)
 &\le\sum_{a,s^\prime}\lambda_{a,s^\prime}
       \E\exp\!\left(\frac n2Z_{s,a,s^\prime}^2\right)
 \le\frac53.
\end{align*}
Thus, for
\[
 \gD_{n,t}=\left\{D_n\le\sqrt2\,|\gS|\sqrt{t/n}\right\},
\]
Markov's inequality and a union bound yield
\begin{equation*}
 \PB(\gD_{n,t}^{\,c})\le\frac53|\gS|e^{-t}.
\end{equation*}

Next, put
\[
 a_{\gI}=1+\log\frac{3|\gI|}{|\gS|},
 \qquad u=t+\log\frac{3|\gI|}{|\gS|}.
\]
Since $t\ge1$ and $|\gI|\ge|\gS|$, we have $1\le u\le a_{\gI}t$. Define
\[
 \gR_{n,t}=\left\{
 \max_{(s,a,s^\prime)\in\gI}
 \abs{\what P(s^\prime\mid s,a)-P(s^\prime\mid s,a)}
 \le\sqrt{\frac{u}{2n}}\right\}.
\]
Hoeffding's inequality gives
\begin{equation*}
 \PB(\gR_{n,t}^{\,c})\le2|\gI|e^{-u}.
\end{equation*}
On $\gR_{n,t}$,
\[
 r_n\le\frac1{p_*}\sqrt{\frac{a_{\gI}t}{2n}}.
\]
Choose
\begin{equation*}
 c_\gM^*=\min\left\{\frac{\varepsilon_0^2}{2|\gS|^2},\frac{p_*^2}{2a_{\gI}}\right\}>0.
\end{equation*}
For $1\le t\le c_\gM^*n$, we have $D_n\le\varepsilon_0$ on $\gD_{n,t}$, and on $\gR_{n,t}$,
\[
 r_n\le\frac12,\qquad
 N_{s,a,s^\prime}\ge\frac{np_*}2,\qquad
 \what P(s^\prime\mid s,a)\le\frac32P(s^\prime\mid s,a)
 \quad((s,a,s^\prime)\in\gI).
\]

Let $C_R=\max_{s,a}C_{F_{s,a}}$, with $C_{F_{s,a}}$ from Lemma~\ref{lem:weighted-ecdf-hp}. Conditional on all successor-state labels, the rewards in each nonempty stratum are independent with CDF $F_{s,a}$. Applying that lemma with sample size $N_{s,a,s^\prime}$ and parameter $u$ to every stratum gives an event $\gB_{n,t}$ such that
\begin{equation*}
 \PB(\gR_{n,t}\cap\gB_{n,t}^{\,c})\le|\gI|e^{-u}
\end{equation*}
and, on $\gR_{n,t}\cap\gB_{n,t}$,
\begin{align*}
 |\what F_{s,a,s^\prime}(r)-F_{s,a}(r)|
 &\le C_R\left\{\sqrt{\frac{u\,p_{s,a}(r)}{N_{s,a,s^\prime}}}
                +\frac{u}{N_{s,a,s^\prime}}\right\}\\
 &\le C_R\left\{\sqrt{\frac{2a_{\gI}t\,p_{s,a}(r)}{np_*}}
                +\frac{2a_{\gI}t}{np_*}\right\}
\end{align*}
for all $(s,a,s^\prime)\in\gI$ and $r\in[0,1]$. Both CDFs agree outside $[0,1]$, so, with $p_{s,a}$ extended by zero, the last bound holds for every $r\in\RB$:
\begin{equation}\label{eq:conditional-reward-bernstein}
 |\what F_{s,a,s^\prime}(r)-F_{s,a}(r)|
 \le C_R\left\{\sqrt{\frac{2a_{\gI}t\,p_{s,a}(r)}{np_*}}
       +\frac{2a_{\gI}t}{np_*}\right\}.
\end{equation}

Let $\gE_{n,t}$ be the intersection of $\gD_{n,t}$, $\gR_{n,t}$, $\gB_{n,t}$ and the probability-one event that $P(s^\prime\mid s,a)=0$ implies $\what P(s^\prime\mid s,a)=0$. Then
\begin{align*}
 \PB(\gE_{n,t}^{\,c})
 &\le\frac53|\gS|e^{-t}+3|\gI|e^{-u}
 =\frac83|\gS|e^{-t}\le4|\gS|e^{-t}.
\end{align*}

Fix $m$, $s$, $\btheta\in\gU_m(\rho)$ and $x\in[0,B]$, and write $y_{s^\prime,j}=x-\gamma\theta(s^\prime,j)$. On $\gE_{n,t}$, the Bellman CDFs satisfy
\begin{align*}
 \what F_{s,\btheta}^{(n)}(x)-F_{s,\btheta}(x)
 &=\sum_{a,s^\prime,j}\frac{\pi(a\mid s)\what P(s^\prime\mid s,a)}m
       [\what F_{s,a,s^\prime}(y_{s^\prime,j})-F_{s,a}(y_{s^\prime,j})]\\
 &\quad+\sum_{a,s^\prime,j}\frac{\pi(a\mid s)}m
       [\what P(s^\prime\mid s,a)-P(s^\prime\mid s,a)]F_{s,a}(y_{s^\prime,j})\\
 &\eqqcolon E_R(x)+E_P(x).
\end{align*}
Since the empirical mixture weights sum to one, \eqref{eq:conditional-reward-bernstein} and Jensen's inequality yield
\begin{align*}
 |E_R(x)|
 &\le C_R\sqrt{\frac{2a_{\gI}t}{np_*}}
      \sum_{a,s^\prime,j}\frac{\pi(a\mid s)\what P(s^\prime\mid s,a)}m
        \sqrt{p_{s,a}(y_{s^\prime,j})}
       +\frac{2a_{\gI}C_R}{p_*}\frac tn\\
 &\le C_R\sqrt{\frac{2a_{\gI}t}{np_*}}
       \left\{\sum_{a,s^\prime,j}
         \frac{\pi(a\mid s)\what P(s^\prime\mid s,a)}m
         p_{s,a}(y_{s^\prime,j})\right\}^{1/2}
       +\frac{2a_{\gI}C_R}{p_*}\frac tn\\
 &\le C_R\sqrt{\frac{3a_{\gI}}{p_*}}
           \sqrt{\frac{t f_{s,\btheta}(x)}n}
       +\frac{2a_{\gI}C_R}{p_*}\frac tn.
\end{align*}
The last step uses $\what P(s^\prime\mid s,a)\le(3/2)P(s^\prime\mid s,a)$ on $\gR_{n,t}$.

For the second term, put
\[
 w_{a,s^\prime,j}=\frac{\pi(a\mid s)P(s^\prime\mid s,a)}m,
 \qquad
 \what w_{a,s^\prime,j}=\frac{\pi(a\mid s)\what P(s^\prime\mid s,a)}m.
\]
Then
\[
 |\what w_{a,s^\prime,j}-w_{a,s^\prime,j}|\le r_n w_{a,s^\prime,j},
 \qquad \sum_{a,s^\prime,j}(\what w_{a,s^\prime,j}-w_{a,s^\prime,j})=0.
\]
Applying these inequalities first to the CDFs and then to their complements gives
\begin{align*}
 |E_P(x)|
 &\le r_n\sum_{a,s^\prime,j}w_{a,s^\prime,j}F_{s,a}(y_{s^\prime,j})
   =r_nF_{s,\btheta}(x),\\
 |E_P(x)|
 &\le r_n\sum_{a,s^\prime,j}w_{a,s^\prime,j}[1-F_{s,a}(y_{s^\prime,j})]
   =r_n[1-F_{s,\btheta}(x)].
\end{align*}
Write $c_{\mathrm{int}}=c_{\gM,\ell}$ and $C_{\mathrm{cdf}}=\max\{B,c_{\mathrm{int}}^{-1}\}$. Boundary monotonicity \eqref{eq:boundary-mono} and the interior density lower bound imply
\[
 \min\{F_{s,\btheta}(x),1-F_{s,\btheta}(x)\}
 \le C_{\mathrm{cdf}}f_{s,\btheta}(x)
 \le C_{\mathrm{cdf}}\sqrt{C_0 f_{s,\btheta}(x)}.
\]
Consequently,
\[
 |E_P(x)|\le r_n C_{\mathrm{cdf}}\sqrt{C_0 f_{s,\btheta}(x)}
 \le\frac{C_{\mathrm{cdf}}}{p_*}\sqrt{\frac{a_{\gI}C_0}{2}}
       \sqrt{\frac{t f_{s,\btheta}(x)}n}.
\]
Combining the two estimates proves \eqref{eq:joint-bernstein-target} with
\[
 K_\gM=\max\left\{
 C_R\sqrt{\frac{3a_{\gI}}{p_*}}
 +\frac{C_{\mathrm{cdf}}}{p_*}\sqrt{\frac{a_{\gI}C_0}{2}},
 \frac{2a_{\gI}C_R}{p_*}\right\}.
\]
All events and constants are independent of $m$ and $\btheta$, so the bound holds simultaneously over the stated neighborhood.
\end{proof}

\subsection{Proof of Lemma~\ref{lem:A4-Winf-one-step}}\label{app:proof-A4-Winf-group}
\phantomsection\label{app:proof-A4-Winf-one-step}
\begin{proof}[Proof of Lemma~\ref{lem:A4-Winf-one-step}]
Fix $(s,a,s^\prime)$ with $N_{s,a,s^\prime}>0$.
For $u\in(0,1)$, let
$x=\what F_{s,a,s^\prime}^{-1}(u)$. By the
definition of the quantile function,
\[
 \what F_{s,a,s^\prime}(x-)
 \le u\le \what F_{s,a,s^\prime}(x).
\]
Since
$\|\what F_{s,a,s^\prime}-F_{s,a}\|_\infty
\le d_n$ and $F_{s,a}$ is continuous,
\[
 u-d_n\le F_{s,a}(x)\le u+d_n.
\]
By Assumption~\ref{ass:A1}, $F_{s,a}$
is strictly increasing on $[0,1]$.
Hence, by the definition of $\omega_R$,
\[
 \abs{\what F_{s,a,s^\prime}^{-1}(u)-F_{s,a}^{-1}(u)}=\abs{F_{s,a}^{-1}(F_{s,a}(x))-F_{s,a}^{-1}(u)}\le\omega_R(d_n).
\]
Coupling the reward variables by these quantile couplings while keeping the action, successor state and continuation atom fixed therefore yields
\begin{equation}\label{eq:A4-Winf-reward-one-step}
 \Wbar_\infty(\bPi_m\what{\gT}_n^\pi\Beta_m,\bPi_m\wtilde{\gT}_n^\pi\Beta_m)\leq \Wbar_\infty(\what{\gT}_n^\pi\Beta_m,\wtilde{\gT}_n^\pi\Beta_m)\le\omega_R(d_n).
\end{equation}

It remains to compare the hybrid and population transition weights. Fix a state and write $F=F_{m,s}$ for the population Bellman-target CDF and $\wtilde F$ for its hybrid version. The proof of Lemma~\ref{lem:A3-one-step} gives
\begin{equation*}
 \abs{\wtilde F(x)-F(x)}\le r_n\min\{F(x),1-F(x)\}.
\end{equation*}
Lemma~B.1 in~\citet{cheng2026statistical} shows that there is a constant $c_\gM>0$, independent of $m,s,i$, such that
\[
 \abs{F(\theta_m(s,i)+h)-\tau_i}\ge c_\gM\tau_i(1-\tau_i)|h|,\qquad 0<|h|\le B.
\]
For $0<r_n\le1/8$, fix $u=\tau_i$ and set
\[
 u_-=u-4r_nu(1-u),\qquad u_+=u+4r_nu(1-u).
\]
The same argument as in Appendix~\ref{app:proof-A3-group} gives
\[
 F^{-1}(u_-)\le\wtilde F^{-1}(u)\le F^{-1}(u_+).
\]
Take $h=F^{-1}(u_\pm)-F^{-1}(u)$ and we have
\[
 \abs{F^{-1}(u_\pm)-F^{-1}(u)}\le\frac{|u_\pm-u|}{c_\gM u(1-u)}=\frac{4r_n}{c_\gM}.
\]
Taking the maximum over states and midpoint levels gives
\begin{equation}\label{eq:A4-Winf-transition-one-step}
 \Wbar_\infty(\bPi_m\wtilde{\gT}_n^\pi\Beta_m,
                 \bPi_m\gT^\pi\Beta_m)
 \le K_\gM r_n.
\end{equation}
For $r_n>1/8$, use the support bound $B\le8Br_n$, and for $r_n=0$, the two targets coincide. Thus \eqref{eq:A4-Winf-transition-one-step} holds for all $r_n\ge0$ after increasing $K_\gM$.

Combining \eqref{eq:A4-Winf-reward-one-step} and \eqref{eq:A4-Winf-transition-one-step} gives
\begin{equation}\label{eq:A4-Winf-primitive-bound}
 \Delta_n^\infty\coloneq \sup_{m\ge1}
 \norm{\what\Phi_{m,n}(\btheta_m)-\Phi_m(\btheta_m)}_\infty
 \le K_{\mathrm{step}}[\omega_R(d_n)+r_n].
\end{equation}
It remains to express this bound in terms of $h_{n,\delta}$. By Lemma~\ref{lem:primitive-conc}, with probability at least $1-\delta$,
\begin{equation}\label{eq:Winf-primitive-event}
 d_n+r_n\le\frac5{p_*}
 \sqrt{\frac{\log(2|\gI|/\delta)}{2n}}.
\end{equation}
Let
\[
 L_{\mathrm{tab}}=\max\left\{1,
 \frac{\log(2|\gI|)}{\log(4|\gS|)}\right\},
 \qquad
 J=\left\lceil\max\left\{1,
 \frac5{16\sqrt2}\sqrt{L_{\mathrm{tab}}}\right\}\right\rceil.
\]
Then $\log(2|\gI|/\delta)\le L_{\mathrm{tab}}\log(4|\gS|/\delta)$. On the event in \eqref{eq:Winf-primitive-event},
\[
 d_n+r_n\le Jg_{n,\delta},
 \qquad
 g_{n,\delta}=\frac{16}{p_*}
 \sqrt{\frac{\log(4|\gS|/\delta)}n}.
\]
For any integer $J\ge1$, we know that $\omega_R(\min\{1,Jh\})\le J\omega_R(h)$ for $0<h\le1$.
If $g_{n,\delta}<1$, then $h_{n,\delta}=g_{n,\delta}$, $d_n\le\min\{1,Jh_{n,\delta}\}$, and $r_n\le Jh_{n,\delta}$. Equation~\eqref{eq:A4-Winf-primitive-bound} therefore gives
\[
 \Delta_n^\infty\le JK_{\mathrm{step}}
 [\omega_R(h_{n,\delta})+h_{n,\delta}].
\]
If $g_{n,\delta}\ge1$, then $h_{n,\delta}=1$ and
$\Delta_n^\infty\le B=(B/2)[\omega_R(1)+1]$.
Taking $K_\gM=\max\{JK_{\mathrm{step}},B/2\}$ proves \eqref{eq:A4-Winf-one-step}.
\end{proof}

\subsection{Proofs of Lemmas~\ref{lem:local-decision}--\ref{lem:local-LF}}\label{app:proof-local-minimax-group}
For the proofs below, define
\[
 [\bH_Q(\btheta)]_{s,i}=F_{[\gT_Q^\pi\Beta_{\btheta}](s)}(\theta(s,i)),
 \qquad
 [\bH_n(\btheta)]_{s,i}=\what F_{s,\btheta}^{(n)}(\theta(s,i)).
\]
Write $[\btau_m]_{s,i}=\tau_i$, so that $\bH_{Q_0}(\btheta_m)=\btau_m$.

Differentiating at $\btheta_m$ gives
\[
 \nabla_{\btheta}\bH_{Q_0}(\btheta_m)=\bG_m.
\]
To establish invertibility, define
\[
 \bD_m=\operatorname{diag}\{f_{m,s}(\theta_m(s,i))\}_{s,i},\qquad\bK_m=\gamma^{-1}(\bI-\bD_m^{-1}\bG_m).
\]
By the definition of $\bG_m$,
\[
 \bK_m\ge0,\qquad\bK_m\mathbf1=\mathbf1,\qquad\bG_m=\bD_m(\bI-\gamma\bK_m),
\]
where $\mathbf1\in\RB^d$ is the all-ones vector. Since $\bD_m$ has strictly positive diagonal entries and $\gamma<1$, $\bG_m$ is invertible.

\phantomsection\label{app:proof-local-decision}
\begin{proof}[Proof of Lemma~\ref{lem:local-decision}]
Monotonicity of the quantile function gives ordering. For $p<\infty$, Jensen's inequality on each probability cell gives
\[
 \frac1m|a_i(\mu)-\theta_i|^p
 \le\int_{(i-1)/m}^{i/m}|F_\mu^{-1}(u)-\theta_i|^p\,\rd u.
\]
Summing proves the claim. For $p=\infty$, we have
\[
\begin{aligned}
 \abs{a_i(\mu)-\theta_i}&=\left|m\int_{(i-1)/m}^{i/m}(F_\mu^{-1}(u)-\theta_i)\,\rd u\right|\\
 &\le\sup_{u\in((i-1)/m,i/m)}\abs{F_\mu^{-1}(u)-\theta_i}.
\end{aligned}
\]
Taking the maximum over $i$ yields
\[
 W_\infty\left(\gA_m\mu,\frac1m\sum_i\delta_{\theta_i}\right)\le W_\infty\left(\mu,\frac1m\sum_i\delta_{\theta_i}\right)
\]
and the conclusion follows. 
\end{proof}

\phantomsection\label{app:proof-joint-local-linearization}
\begin{proof}[Proof of Lemma~\ref{lem:joint-local-linearization}]
We know that
\[
 \bH_Q(\btheta_m(Q))=\btau_m,\ \bH_n(\btheta_m^{(n)})=\btau_m+\bm\epsilon _n,
\]
where $\|\bm\epsilon _n\|_\infty=O(n^{-1})$.
Let $\btheta_Q=\btheta_m(Q)$ and $\btheta_0=\btheta_m(Q_0)$. Lemma~\ref{lem:local-moments} gives $\sqrt n\|\btheta_Q-\btheta_0\|_\infty\le L_{m,Q_0}c$ and, for each $r\ge1$,
\[
 \sup_{Q\in\gN_n(c;Q_0)}
 \PB_Q\brc{\sqrt n\norm{\btheta_m^{(n)}-\btheta_0}_\infty>M}\eqqcolon\sup_{Q\in\gN_n(c;Q_0)}\PB_Q \brc{\gE_{n,M}^c}
 \le C_{m,c,r,Q_0}M^{-r}.
\]
Fix $M>L_{m,Q_0}c$ and we work on the event $\gE_{n,M}$.

At the center $Q_0$, Assumption~\ref{ass:A2} and continuity of the reward densities imply
\[
 \bH_{Q_0}(\btheta_0+\bu)=\btau_m+\bG_m\bu+\bR_0(\bu),
 \qquad \|\bR_0(\bu)\|_\infty\le\|\bu\|_\infty\omega_m(\|\bu\|_\infty),
\]
where $\omega_m(t)\to0$ as $t\downarrow0$. Subtracting the two estimating equations gives the exact decomposition
\begin{align*}
 \bG_m\sqrt n(\btheta_m^{(n)}-\btheta_Q)
 ={}&-\sqrt n(\bH_n-\bH_Q)(\btheta_0)\\
 &+\sqrt n\bm\epsilon _n\\
 &-\sqrt n\{(\bH_n-\bH_Q)(\btheta_m^{(n)})-(\bH_n-\bH_Q)(\btheta_0)\}\\
 &-\sqrt n\{(\bH_Q-\bH_{Q_0})(\btheta_m^{(n)})-(\bH_Q-\bH_{Q_0})(\btheta_Q)\}\\
 &-\sqrt n\{\bR_0(\btheta_m^{(n)}-\btheta_0)-\bR_0(\btheta_Q-\btheta_0)\}.
\end{align*}
Now we bound the four terms. The last term satisfies
\begin{align*}
&\sqrt n\norm{\bR_0(\btheta_m^{(n)}-\btheta_0)-\bR_0(\btheta_Q-\btheta_0)}_\infty\\
\le&\sqrt n\prn{\|\btheta_m^{(n)}-\btheta_0\|_\infty+\|\btheta_Q-\btheta_0\|_\infty}\omega_m(M/\sqrt n)\\
\le& 2M\omega_m(M/\sqrt n)\to 0.
\end{align*}
For the second term, Lemma~\ref{lem:local-increments} with $\delta=M/\sqrt n$ and $d_H(Q,Q_0)\le c/\sqrt n$ gives
\begin{align*}
&\sup_{Q\in\gN_n(c;Q_0)}\E_Q\brk{\sqrt n\norm{\prn{\bH_n-\bH_Q}\prn{\btheta_m^{(n)}}-\prn{\bH_n-\bH_Q}\prn{\btheta_0}}_\infty
\ind\{\mathcal E_{n,M}\}}\\
\le&C_{m,Q_0}\sqrt{\frac{M}{\sqrt n}+\frac{c^2}{n}}\to0.
\end{align*}
Hence the second term converges to zero in probability uniformly over $Q\in\gN_n(c;Q_0)$ on $\mathcal E_{n,M}$, by Markov's inequality.

For the third term, both $\btheta_m^{(n)}$ and $\btheta_Q$ belong to the radius-$M/\sqrt n$ neighborhood of $\btheta_0$ on $\mathcal E_{n,M}$.
Thus Lemma~\ref{lem:local-increments} yields
\begin{align*}
&\sqrt n\norm{\prn{\bH_Q-\bH_{Q_0}}\prn{\btheta_m^{(n)}}-\prn{\bH_Q-\bH_{Q_0}}\prn{\btheta_Q}}_\infty\\
\le&2\sqrt n\sup_{\norm{\btheta-\btheta_0}_\infty\le M/\sqrt n}
\norm{\prn{\bH_Q-\bH_{Q_0}}\prn{\btheta}-\prn{\bH_Q-\bH_{Q_0}}\prn{\btheta_0}}_\infty\\
\le&2C_{m,Q_0}c\sqrt{\frac{M}{\sqrt n}+\frac{c^2}{n}}\to0.
\end{align*}
Finally $\sqrt n\|\bm\epsilon _n\|_\infty=O(n^{-1/2})$. Thus all remainders vanish for fixed $M$. Let $M\to\infty$ and we obtain
\begin{equation*}
 \bU_{n,Q}=-\bG_m^{-1}\frac1{\sqrt n}\sum_{k=1}^n\sum_b\brc{\bY_{m,b}(X_{k,b})-\E_{Q_b}\bY_{m,b}}+\bm\epsilon _{n,Q},
\end{equation*}
where for every $\varepsilon>0$,
\[
 \lim_{n\to\infty}\sup_{Q\in\gN_n(c;Q_0)}\PB_Q\prn{\norm{\bm\epsilon _{n,Q}}_\infty>\varepsilon}=0.
\]

Fix any sequence $Q_n\in\gN_n(c;Q_0)$. 
Their covariance matrices satisfy
\[
 \begin{aligned}
 &\norm{\sum_b\operatorname{Cov}_{Q_{n,b}}(\bY_{m,b})-\bSigma_m}_\infty\\
 \le& C_m\max_b\TV(Q_{n,b},Q_{0,b})\le \frac{C_m c}{\sqrt n}\to0.
 \end{aligned}
\]
Thus the Lindeberg central limit theorem gives
\[
 \frac1{\sqrt n}\sum_{k=1}^n\sum_b
 \brc{\bY_{m,b}(X_{k,b})
 -\E_{Q_{n,b}}\bY_{m,b}}
 \cd\normal(0,\bSigma_m).
\]
Since $d_{\mathrm{BL}}$ metrizes weak convergence
on $\RB^d$, this implies
\[
 d_{\mathrm{BL}}\prn{\mathcal L_{Q_n}(\bU_{n,Q_n}-\bm\epsilon_{n,Q_n}),\normal(0,\bV_m)}\to0.
\]
As the sequence $Q_n$ was arbitrary, we obtain
\[
 \sup_{Q\in\gN_n(c;Q_0)}d_{\mathrm{BL}}\prn{\mathcal L_Q(\bU_{n,Q}-\bm\epsilon_{n,Q}),\normal(0,\bV_m)}\to0.
\]

It remains to control the effect of the remainder.
For any function $h$ in the definition of
$d_{\mathrm{BL}}$ and any $\varepsilon>0$,
\begin{align*}
&\abs{\E_Q h(\bU_{n,Q})
 -\E_Q h(\bU_{n,Q}-\bm\epsilon_{n,Q})}\\
\le&
 \E_Q\min\brc{2,\sqrt d\norm{\bm\epsilon_{n,Q}}_\infty}\\
\le&
 \sqrt d\,\varepsilon
 +2\PB_Q\prn{\norm{\bm\epsilon_{n,Q}}_\infty>\varepsilon}.
\end{align*}
Consequently,
\begin{align*}
&\sup_{Q\in\gN_n(c;Q_0)}
 d_{\mathrm{BL}}\prn{
 \mathcal L_Q(\bU_{n,Q}),\normal(0,\bV_m)}\\
\le&
 \sup_{Q\in\gN_n(c;Q_0)}
 d_{\mathrm{BL}}\prn{
 \mathcal L_Q(\bU_{n,Q}-\bm\epsilon_{n,Q}),
 \normal(0,\bV_m)}\\
&+\sqrt d\,\varepsilon
 +2\sup_{Q\in\gN_n(c;Q_0)}
 \PB_Q\prn{\norm{\bm\epsilon_{n,Q}}_\infty>\varepsilon}.
\end{align*}
Taking $n\to\infty$ and then
$\varepsilon\downarrow0$ proves the asserted
uniform bounded Lipschitz convergence.

Finally, Lemma~\ref{lem:local-moments} gives uniform $q$-moment bounds for $\bU_{n,Q}$ for every $q>1$.
Since $\ell_{m,p}(\bz)\le\|\bz\|_\infty$, these bounds imply the asserted uniform integrability for every $1\le p\le\infty$.
\end{proof}

\phantomsection\label{app:proof-local-LF}
\begin{proof}[Proof of Lemma~\ref{lem:local-LF}]
Let $\HB=\bigoplus_bL_0^2(Q_{0,b})$, with inner product $\langle h,g\rangle_{\HB}=\sum_b\E_{Q_{0,b}}[h_bg_b]$.
For bounded $h=(h_b)_b\in\HB$, consider the parametric submodel
\[
 \rd Q_{\epsilon,b}=(1+\epsilon h_b)\,\rd Q_{0,b},
\]
and write $\btheta_\epsilon=\btheta_m(Q_\epsilon)$.
Differentiating $\bH_{Q_\epsilon}(\btheta_\epsilon)=\btau_m$ at $\epsilon=0$ gives
\begin{equation}\label{eq:joint-path-derivative}
 \dot\btheta_m[h]\coloneq\left.\frac{\rd}{\rd\epsilon}\btheta_m(Q_\epsilon)\right|_{\epsilon=0}=-\bG_m^{-1}\sum_b\E_{Q_{0,b}}[\bY_{m,b}h_b]=\sum_b\E_{Q_{0,b}}[\bphi_{m,b}h_b].
\end{equation}
Bounded mean-zero scores are dense in $\HB$, so $\bphi_m=(\bphi_{m,b})_b$ is the canonical gradient and each of its components is bounded at fixed $m$.

Let $h^1,\ldots,h^r$ be an orthonormal basis of $\operatorname{span}\{(\phi_{m,b}(s,i))_b:(s,i)\in\gS\times[m]\}\subset\HB$, then every $h^k_b$ is bounded. Put
\begin{equation*}
 \rd Q_{\bt,b}
 =\left(1+\sum_{k=1}^r t_k h_b^k\right)\rd Q_{0,b}
\end{equation*}
for $\bt$ in a sufficiently small neighbourhood of zero. Define
\[
 A_{(u,i),k}=\sum_b\E_{Q_{0,b}}[\phi_{m,b}(u,i)h_b^k].
\]
Expansion of the coordinate gradients in their orthonormal basis gives $\bA\bA^\top=\bV_m$. 
For each $b$ and $k$, the score function is
\[
 \left.\frac{\partial}{\partial t_k}
 \log\frac{\rd Q_{\bt,b}}{\rd Q_{0,b}}
 \right|_{\bt=0}=h_b^k.
\]
Thus the Fisher information matrix is $\bI_r$. 
Moreover, \eqref{eq:joint-path-derivative} gives
\[
 \left.\frac{\partial}{\partial t_k}
 \btheta_m(Q_{\bt})\right|_{\bt=0}
 =\sum_b\E_{Q_{0,b}}[\bphi_{m,b}h_b^k]
 =\bA_{\cdot,k}.
\]

To study the local likelihoods, put
\[
 \Delta_{n,k}=\frac1{\sqrt n}\sum_{j=1}^n\sum_b h_b^k(X_{j,b}),\quad \boldsymbol{\Delta}_n\coloneq(\Delta_{n,1},\ldots,\Delta_{n,r})^\top.
\]
Taylor expansion of $\log(1+x)$ for the bounded scores and the law of large numbers give, uniformly on every fixed compact set of $\bt$,
\begin{equation*}
 \log\frac{\rd\PB_{Q_{\bt/\sqrt n}}^{(n)}}{\rd\PB_{Q_0}^{(n)}}
 =\bt^\top\boldsymbol{\Delta}_n-\frac12\|\bt\|_2^2+o_{\PB_{Q_0}}(1),
\qquad\boldsymbol{\Delta}_n\cd\normal(0,\bI_r).
\end{equation*}
The target derivative above and the expansion of the square roots of the densities also give
\begin{equation*}
\begin{aligned}
 \sqrt n\{\btheta_m(Q_{\bt/\sqrt n})-\btheta_m(Q_0)\}
 &=\bA\bt+o(1),\\
 n\,d_H(Q_{\bt/\sqrt n},Q_0)^2
 &=\tfrac14\|\bt\|_2^2+o(1),
\end{aligned}
\end{equation*}
uniformly on compact sets. For a finite set $T\subset\RB^r$, choose $c>\frac12\max_{\bt\in T}\|\bt\|_2$. Then all $Q_{\bt/\sqrt n}$ with $\bt\in T$ belong to $\gN_n(c;Q_0)$ for sufficiently large $n$, as required.
\end{proof}

\subsection{Proofs of Lemmas~\ref{lem:complexity-positive}--\ref{lem:hard-cdf-variance}}\label{app:proof-hard-instance-group}
\phantomsection\label{app:proof-complexity-positive}
\begin{proof}[Proof of Lemma~\ref{lem:complexity-positive}]
Let $\mu_{s,a}=\E R^{(s,a)}$ and choose the bounded
score direction
\[
 h_{s,a}(r,s^\prime)=r-\mu_{s,a}.
\]
Then $0<\|h\|_{\HB}^2
=\sum_{s,a}\var(R^{(s,a)})<\infty$.
Define
\[
 b_{s,a}(x)=-\int_0^x(r-\mu_{s,a})p_{s,a}(r)\,\rd r
\]
on $[0,1]$, and extend it by zero outside. Since
\[
 b_{s,a}'(x)=(\mu_{s,a}-x)p_{s,a}(x),\qquad b_{s,a}(0)=b_{s,a}(1)=0,
\]
we have $b_{s,a}(x)>0$ for $x\in(0,1)$, and $b_{s,a}$ is globally $C_0$-Lipschitz.

Define the vector $\bu_m=(u_m(s,i))_{s,i}$ by
\[
 u_m(s,i)=\sum_{a,s^\prime}\pi(a\mid s)P(s^\prime\mid s,a)\int b_{s,a}(\theta_m(s,i)-\gamma y)\,\eta_m(s^\prime)(\rd y).
\]
By \eqref{eq:joint-path-derivative},
\begin{equation*}
 \dot\btheta_m[h]=(\bI-\gamma\bK_m)^{-1}\bD_m^{-1}\bu_m\ge\bD_m^{-1}\bu_m\ge0,
\end{equation*}
where the inequalities are coordinatewise. Since $\bD_m\le C_0\bI$, writing $L_s(\bz)=m^{-1}\sum_i z(s,i)$ gives
\[
 L_s(\dot\btheta_m[h])
 \ge\frac1{C_0m}\sum_{i=1}^m u_m(s,i).
\]

For each $s$, define
\[
 \beta_s=\sum_{a,s^\prime}\pi(a\mid s)P(s^\prime\mid s,a)\iint b_{s,a}(x-\gamma y)\,\eta^\pi(s)(\rd x)\eta^\pi(s^\prime)(\rd y).
\]
By Proposition~\ref{prop:return_density} and the positivity of $b_{s,a}$ on $(0,1)$, we have $\beta_s>0$. Moreover, the Lipschitz continuity of $b_{s,a}$ and \eqref{eq:basic-approx} give
\[
 \abs{\frac1m\sum_{i=1}^m u_m(s,i)-\beta_s}
 \le C_0(1+\gamma)
 \Wbar_1(\Beta_m,\Beta^\pi)
 =O(m^{-1}).
\]
Consequently,
\[
 \liminf_{m\to\infty}L_s(\dot\btheta_m[h])
 \ge\frac{\beta_s}{C_0}>0.
\]
For every fixed $m$, the positivity of
$f_{m,s}(\theta_m(s,i))$ implies
$u_m(s,i)>0$. Hence, fixing any state $s_0$,
\[
 a_{Q_0}\coloneq
 \inf_{m\ge1}L_{s_0}(\dot\btheta_m[h])>0.
\]
Therefore, 
\[
\begin{aligned}
 a_{Q_0}&\le L_{s_0}(\dot\btheta_m[h])\\
 &=\sum_b\E_{Q_{0,b}}[L_{s_0}(\bphi_{m,b})h_b]\\
 &\le\|h\|_{\HB}\sqrt{\var(L_{s_0}(\bZ_m))}.
\end{aligned}
\]
Finally, since $|L_{s_0}(\bz)|\le\ell_{m,1}(\bz)$,
\[
 \mathfrak C_{m,1}(Q_0)\ge\E|L_{s_0}(\bZ_m)|\ge\sqrt{\frac2\pi}\frac{a_{Q_0}}{\|h\|_{\HB}}>0,
\]
and Lemma~\ref{lem:complexity-positive} holds. 
\end{proof}

\phantomsection\label{app:proof-uniform-boundary}
\begin{proof}[Proof of Lemma~\ref{lem:uniform-boundary}]
Fix $m$ and write $\theta_i=\theta_m(i)$. Let
\[
 A_i=\{j:0<\theta_i-\gamma\theta_j<1\},\qquad
 a_i=|A_i|,\qquad b_i=\#\{j:\theta_i-\gamma\theta_j\ge1\}.
\]
The uniform-reward CDF equation is
\begin{equation*}
 a_i\theta_i-\gamma\sum_{j\in A_i}\theta_j=m\tau_i-b_i.
\end{equation*}
If $a_i=0$, its CDF value would be $b_i/m$, which cannot equal $(i-1/2)/m$. Thus $a_i\ge1$. 
For the fixed sets $A_i$, define
\[
 C_{ij}=\ind\{j\in A_i\},\qquad M(z)=\operatorname{diag}(a_i)-zC.
\]
Since $\operatorname{diag}(a_i)^{-1}C$ is row-stochastic, $M(\gamma)$ is invertible. Writing $v_i=i-1/2-b_i$, we have
\[
 \btheta=M(\gamma)^{-1}\bv.
\]
By Cramer's rule, each $\theta_i$ can be written as $\Theta_i(\gamma)$ for some rational function $\Theta_i$ with rational coefficients.

Suppose that $\theta_i-\gamma\theta_j=c\in\{0,1\}$. Then
\[
\begin{aligned}
    \Theta_i(\gamma)-\gamma\Theta_j(\gamma)-c&=0,\\
    \Theta_i(0)-c=\frac{i-1/2-b_i}{a_i}-c&\ne0.
\end{aligned} 
\]
This shows that $\gamma$ is a zero of a polynomial with rational coefficients, which contradicts the transcendence of $\gamma$.
\end{proof}

\phantomsection\label{app:proof-hard-rewards}
\begin{proof}[Proof of Lemma~\ref{lem:hard-rewards}]

We first construct a $C^1$ density $p_s^0$, called the shelf density. For every sufficiently large $j$, prescribe
\[
 g(r)=\epsilon_j+\frac{4\epsilon_j}{3x_j}(r-x_j/4),\qquad r\in[x_j/4,x_j].
\]
Since $x_{j+1}=x_j/8$ and $\epsilon_{j+1}/\epsilon_j\to0$, the adjacent segments can be joined by strictly increasing $C^1$ interpolations on $I_j=[x_j/8,x_j/4]$, matching the endpoint derivatives and satisfying
\[
 0<g'(r)\le C\epsilon_j/x_j,\qquad r\in I_j.
\]
Indeed, the required increase is $\epsilon_j-2\epsilon_{j+1}\asymp\epsilon_j$ over an interval of length $x_j/8$, and both endpoint derivatives are $O(\epsilon_j/x_j)$.

Since $\epsilon_j/x_j\to0$, the resulting function extends to zero with $g(0)=g'(0)=0$. Extend it as a positive $C^1$ function on $[0,1]$, decreasing to zero near one, and normalize its integral to obtain $p_s^0$. Then $p_s^0$ is Lipschitz, strictly increasing near zero, and
\[
 p_s^0(r)\asymp\epsilon_j \quad(r\in[x_j/4,x_j]),\qquad \sup_{0\le r\le x_j}p_s^0(r)\le C\epsilon_j.
\]

Set $p_t^0(r)=6r(1-r)$. Lemma~\ref{lem:generic} provides perturbed densities $p_s,p_t$ satisfying Assumptions~\ref{ass:A1},~\ref{ass:A2}, and~\ref{ass:A4}, with
\[
 \frac34p_u^0(r)\le p_u(r)\le\frac54p_u^0(r),
 \qquad u\in\{s,t\},\quad 0\le r\le1.
\]
These inequalities give the asserted density bounds. Fix this MDP and denote it by $Q_*$.
\end{proof}

\phantomsection\label{app:proof-hard-resolutions}
\begin{proof}[Proof of Lemma~\ref{lem:hard-resolutions}]
Apply Lemma~\ref{lem:hard-localization} to the fixed reward laws from Lemma~\ref{lem:hard-rewards}. With the same definitions of $H,H_m,a_j$, it gives
\[
 c\epsilon_jx_j e^{-Cj^2}\le a_j\le C\epsilon_jx_j,
 \qquad q_j\in[x_j/2,x_j].
\]
Since $a_j\to0$ and $m_j$ is the integer nearest to $(2a_j)^{-1}$, we have $m_j\asymp a_j^{-1}$. This proves the two bounds on $m_j$. Taking logarithms and using $x_j=8^{-j}$ and $\epsilon_j=e^{-e^{j^2}}$ yields $\log m_j=e^{j^2}+O(j^2)$.

It remains to bound the probability under $\nu_{m_j}$. For probability measures $\mu,\nu$ and $h>0$, a coupling and Markov's inequality give
\[
 F_\mu(x)\ge F_\nu(x-h)-\frac{W_1(\mu,\nu)}h.
\]
By \eqref{eq:basic-approx} and Lemma~\ref{lem:hard-localization}, for sufficiently large $j$,
\[
 \nu_{m_j}([0,x_j/4])\ge \nu([0,x_j/8])-\frac{C}{m_jx_j}\ge e^{-Cj^2}-C\epsilon_j\geq e^{-2Cj^2},
\]
where the last inequality uses the lower bound on $m_j$. 
\end{proof}

\phantomsection\label{app:proof-hard-cdf-variance}
\begin{proof}[Proof of Lemma~\ref{lem:hard-cdf-variance}]
Since $q_j$ is the first midpoint quantile of the continuous CDF $H_{m_j}$,
\[
 \E g_j(R_s)=H_{m_j}(q_j)=\frac1{2m_j}.
\]
For $r\in[x_j/4,3x_j/8]$, the bound $q_j\ge x_j/2$ and $\gamma=1/2$ imply $(q_j-r)/\gamma\ge x_j/4$. Lemma~\ref{lem:hard-resolutions} therefore gives
\[
 g_j(r)\ge\nu_{m_j}([0,x_j/4])\ge e^{-Cj^2}.
\]
By Lemma~\ref{lem:hard-rewards}, this reward interval has probability at least $c\epsilon_jx_j$. Therefore there exists a sufficiently large constant $C$ such that,
\[
 \E g_j(R_s)^2\ge c\epsilon_jx_j e^{-Cj^2}.
\]
The lower bound on $m_j$ also gives
\[
 \{\E g_j(R_s)\}^2
 \le C\epsilon_j^2x_j^2
 =o\!\left(\epsilon_jx_j e^{-Cj^2}\right).
\]
Therefore,
\begin{equation*}
 \var(g_j(R_s))\ge c\epsilon_jx_j e^{-Cj^2}.
\end{equation*}
Finally, $q_j\le x_j$ and $Z_{m_j}\sim\nu_{m_j}$ is nonnegative. The density $p_s$ is extended by zero outside $[0,1]$, so Lemma~\ref{lem:hard-rewards} implies
\begin{equation*}
 f_{m_j,s}(q_j)=\E p_s(q_j-\gamma Z_{m_j})
 \le C\epsilon_j,
\end{equation*}
and Lemma~\ref{lem:hard-cdf-variance} holds.
\end{proof}

%% file: appendices/basic.tex
We use the stratum notation and errors $d_n,r_n$ from Appendix~\ref{app:proof-A3-group}. For the joint CDFs used in \eqref{eq:Dn-definition}, write
\[
 M_n\coloneq\max_{s,a,s^\prime}\norm{\what J_{s,a,s^\prime}-J_{s,a,s^\prime}}_\infty.
\]

\begin{lemma}\label{lem:primitive-conc}
For every $\delta\in(0,1)$ and every $n\ge1$,
\begin{equation}\label{eq:primitive-explicit-hp}
  \PB\left(
  d_n+r_n\le \frac5{p_*}
  \sqrt{\frac{\log(2|\gI|/\delta)}{2n}}
  \right)\ge1-\delta.
\end{equation}
Moreover,
\begin{equation}\label{eq:primitive-explicit-mean}
  \E(d_n+r_n)\le \frac5{p_*}
  \sqrt{\frac{1+\log(2|\gI|)}{2n}}.
\end{equation}
\end{lemma}

\begin{proof}
For every positive-probability stratum, evaluation at $x=1$ gives
\[
  \abs{\what P(s^\prime\mid s,a)-P(s^\prime\mid s,a)}\le M_n,
  \qquad r_n\le M_n/p_*.
\]
If $M_n\le p_*/2$, every such stratum is nonempty and
$\what P(s^\prime\mid s,a)\ge P(s^\prime\mid s,a)/2$.
Using $J_{s,a,s^\prime}=P(s^\prime\mid s,a)F_{s,a}$ and
$\what F_{s,a,s^\prime}=\what J_{s,a,s^\prime}/\what P(s^\prime\mid s,a)$,
we obtain
\begin{align*}
 \norm{\what F_{s,a,s^\prime}-F_{s,a}}_\infty
 &\le
 \frac{\norm{\what J_{s,a,s^\prime}-J_{s,a,s^\prime}}_\infty
       +\abs{\what P(s^\prime\mid s,a)-P(s^\prime\mid s,a)}}
      {\what P(s^\prime\mid s,a)}\\
 &\le\frac{4M_n}{P(s^\prime\mid s,a)}
 \le\frac{4M_n}{p_*}.
\end{align*}
If $M_n>p_*/2$, the bound $d_n\le1\le2M_n/p_*$ holds, including for empty strata under the stated convention.
Consequently, on the whole sample space,
\begin{equation}\label{eq:primitive-pathwise}
 d_n+r_n\le\frac{5M_n}{p_*}.
\end{equation}
For each $(s,a,s^\prime)\in\gI$, define
\[
 Y_k^{(s,a,s^\prime)}=
 \begin{cases}
 R_k^{(s,a)},&S_k^{\prime(s,a)}=s^\prime,\\
 2,&S_k^{\prime(s,a)}\ne s^\prime.
 \end{cases}
\]
For $x\in[0,1]$,
\[
 \ind\{Y_k^{(s,a,s^\prime)}\le x\}
 =\ind\{R_k^{(s,a)}\le x,\ S_k^{\prime(s,a)}=s^\prime\}.
\]
Their empirical and population CDFs agree with $\what J_{s,a,s^\prime}$ and $J_{s,a,s^\prime}$ on $[0,1]$. The DKW inequality \citep{massart1990tight} and a union bound over $\gI$ give
\[
 \PB(M_n>x)\le2|\gI|e^{-2nx^2},\qquad x>0.
\]
Combining \eqref{eq:primitive-pathwise} with this tail bound at $x=\sqrt{\log(2|\gI|/\delta)/(2n)}$ proves \eqref{eq:primitive-explicit-hp}. Integrating the tail bound and applying \eqref{eq:primitive-pathwise} gives \eqref{eq:primitive-explicit-mean}.
\end{proof}

%% file: appendices/geometry.tex
\begin{lemma}\label{lem:column}

Under Assumption~\ref{ass:A4}, fix $0<\ell<\min\{\kappa,1/2,B/2\}$ and $0<\rho\le\rho_0(\gM,\ell)$ as in
Lemma~\ref{lem:common-localization}. For every $\zeta>0$, there exists $M_0=M_0(\gM,\ell,\zeta)<\infty$ such that for all $m\ge M_0$, every $\btheta\in\gU_m(\rho)$, every $s,a$, and every shift $b\in[0,B-1]$,
\begin{equation}\label{eq:column-average}
  \frac1m\sum_{i=1}^m
  \frac{p_{s,a}(q_i-b)}{f_{s,\btheta}(q_i)}
  \le1+4C_0\ell+\zeta,
\end{equation}
where $q_i=F_{s,\btheta}^{-1}(\tau_i)$.
\end{lemma}
\begin{proof}
Write $q(v)=F_{s,\btheta}^{-1}(v)$ and split the sum into $q_i<\ell$, $\ell\le q_i\le B-\ell$, and $q_i>B-\ell$. On the left, set $a_i=(i-1)/m$. For $a_i<v\le\tau_i$ and $q_i<\ell$, boundary monotonicity gives
\[
 \frac1{f_{s,\btheta}(q_i)}\le\frac1{f_{s,\btheta}(q(v))}.
\]
Therefore, 
\begin{align*}
    \frac1m\sum_{q_i<\ell}\frac{p_{s,a}(q_i-b)}{f_{s,\btheta}(q_i)}&\leq \frac{C_0}m\sum_{q_i<\ell}\frac{1}{f_{s,\btheta}(q_i)}\\
    &\le2C_0\int_0^{F_{s,\btheta}(\ell)}\frac{\rd v}{f_{s,\btheta}(q(v))}\le2C_0\ell.
\end{align*}
The symmetric argument gives the same bound on the right.

For the interior, abbreviate $c=c_{\gM,\ell}$ and define
\[
 r(x)=\frac{p_{s,a}(x-b)}{f_{s,\btheta}(x)},\quad x\in[\ell,B-\ell],
 \qquad
 h(u)=r(q(u))\ind\{\ell\le q(u)\le B-\ell\}.
\]
Since $f_{s,\btheta}\ge c$ on the interior,
\[
 |r(x)-r(y)|\le\left(\frac1c+\frac{C_0}{c^2}\right)\varpi_R(|x-y|).
\]
Moreover, $|q(u)-q(v)|\le c^{-1}|u-v|$ whenever the two quantiles $q(u)$, $q(v)$ lie in the interior interval. 
Therefore, on a probability cell whose interior does not contain either cut level $F_{s,\btheta}(\ell)$ or $F_{s,\btheta}(B-\ell)$, it follows that, for almost every $u$ in that cell,
\[
 |h(\tau_i)-h(u)|\le\left(\frac1c+\frac{C_0}{c^2}\right)\varpi_R\!\left(\frac1{mc}\right).
\]
There are at most two exceptional cells and their contributions are each at most $C_0/(mc)$, since $h\le C_0/c$. 
Consequently,
\begin{align*}
 &\frac1m\sum_{i:\,\ell\le q_i\le B-\ell}\frac{p_{s,a}(q_i-b)}{f_{s,\btheta}(q_i)}\\
 =&\frac1m\sum_i h(\tau_i)\\
 \leq&\int_0^1h(u)\,\rd u+\abs{\frac1m\sum_i h(\tau_i)-\int_0^1h(u)\,\rd u}\\
 \le&1+\left(\frac1c+\frac{C_0}{c^2}\right)
               \varpi_R\!\left(\frac1{mc}\right)+\frac{2C_0}{mc}.
\end{align*}
The last two terms tend to zero uniformly in the stated inputs and shifts. Choose $M_0$ so that they are at most $\zeta$ for $m\ge M_0$. Adding the two boundary estimates proves \eqref{eq:column-average}.
\end{proof}


\begin{lemma}\label{lem:shifted-envelopes}
Under Assumption~\ref{ass:A4}, let $\ell$ and $\rho_0=\rho_0(\gM,\ell)$ be as in Lemma~\ref{lem:common-localization}. Fix $n,m\ge1$, $0<\rho\le\rho_0$, $s\in\gS$ and $\btheta\in\gU_m(\rho)$, and suppose that for some $z\ge0$ and $0\le\varepsilon<1/4$, for any $x\in I$, 
\[
 \abs{\what F_{s,\btheta}^{(n)}(x)-F_{s,\btheta}(x)}\le z\sqrt{f_{s,\btheta}(x)}+\varepsilon.
\]
For $\varepsilon<u<1-\varepsilon$, set $v_\pm=u\pm\varepsilon$ and $r(u)=F_{s,\btheta}^{-1}(v_+)-F_{s,\btheta}^{-1}(v_-)$. Then there exist nonnegative functions $H_L,H_R$ on $(0,1)$, with $H_L$ nonincreasing and $H_R$ nondecreasing, satisfying
\begin{equation}\label{eq:envelope-integrals}
 \int_0^1[H_{L}(v)+H_{R}(v)]\,\rd v\le14\ell z^2
\end{equation}
and
\begin{equation}\label{eq:shifted-pointwise}
 \abs{\brk{\what F_{s,\btheta}^{(n)}}^{-1}(u)-F_{s,\btheta}^{-1}(u)}^2\le C_{\mathrm{geom}}\bigl[r(u)^2+z^2
  +H_{L}(v_-)+H_{R}(v_+)\bigr],
\end{equation}
where $C_{\mathrm{geom}}$ depends only on $\gM$ and $\ell$. 
\end{lemma}
\begin{proof}
Write
\[
 F=F_{s,\btheta},\qquad G=\what F_{s,\btheta}^{(n)},\qquad
 f=f_{s,\btheta},\qquad q=F^{-1},\qquad \what q=G^{-1}.
\]
By Lemma~\ref{lem:common-localization} and \eqref{eq:boundary-mono}, $f$ is positive on the interior of its support. Thus $q'(v)=1/f(q(v))$ for $0<v<1$. Write $g=q'$. First suppose $z>0$. On the left define
\[
 H_{L,+}(v)=4z^2g(v)\ind\{v<F(\ell)\}.
\]
It is nonincreasing, and the change of variables $v=F(x)$ on the active left support gives
\begin{equation*}
 \int_0^1 H_{L,+}(v)\,\rd v
 =4z^2\int_0^{F(\ell)}g(v)\,\rd v\le4\ell z^2.
\end{equation*}
Now we set $A_z(x)=F(x)+z\sqrt{f(x)}$ on $[0,\ell]$, and let $a_z=A_z^{-1}$ on $(0,A_z(\ell)]$. The map $A_z$ is continuous and nondecreasing; it is strictly increasing on the active left support. Define
\[
 H_{L,-}(v)=\min\left\{\ell^2,\frac{z^2}{f(a_z(v))}\right\}\ind\{0<v\le A_z(\ell)\}.
\]
This envelope is nonincreasing, including the terminal truncation. On an initial zero-density interval, $\rd A_z=0$; define the bounded integrand there by its limiting value. Stieltjes change of variables, followed by $\rd A_z=f\,\rd x+z\,\rd\sqrt f$, gives
\begin{align*}
 \int_0^1H_{L,-}(v)\,\rd v
 &\le\int_0^\ell\min\left\{\ell^2,\frac{z^2}{f(x)}\right\}
                    \rd\bigl(F(x)+z\sqrt{f(x)}\bigr)\notag\\
 &\le\ell z^2+z\int_0^\infty
                 \min\left\{\ell^2,\frac{z^2}{y^2}\right\}\,\rd y
 =3\ell z^2.
\end{align*}
The last integral is $2\ell z$, obtained by splitting at $y=z/\ell$. For the right boundary, apply the same construction to
\[
 F^{\mathrm{ref}}(x)=1-F(B-x),\qquad
 f^{\mathrm{ref}}(x)=f(B-x),\qquad
 G^{\mathrm{ref}}(x)=1-G((B-x)-).
\]
By continuity of $F$,
\[
 |G^{\mathrm{ref}}(x)-F^{\mathrm{ref}}(x)|
 \le z\sqrt{f^{\mathrm{ref}}(x)}+\varepsilon.
\]
If $G(y-)\le u\le G(y)$, then
\[
 G^{\mathrm{ref}}((B-y)-)=1-G(y)\le1-u
 \le1-G(y-)=G^{\mathrm{ref}}(B-y).
\]
Define
\[
 H_{R,+}(v)=H_{L,-}^{\mathrm{ref}}(1-v),\qquad
 H_{R,-}(v)=H_{L,+}^{\mathrm{ref}}(1-v).
\]
They are nondecreasing and have integral bounds $3\ell z^2$ and $4\ell z^2$. Set
\[
 H_L=H_{L,+}+H_{L,-},\qquad H_R=H_{R,+}+H_{R,-}.
\]
Both functions have the asserted monotonicity, and their combined integral is at most $14\ell z^2$.

We now verify Equation~\eqref{eq:shifted-pointwise}. Fix $y$ with $G(y-)\le u\le G(y)$. For $x<y$, $G(x)\le u$, whereas $G(y)\ge u$. Therefore,
\begin{equation}\label{eq:shifted-cdf-relations}
 F(x)-v_+\le z\sqrt{f(x)},
 \qquad F(y)+z\sqrt{f(y)}\ge v_-.
\end{equation}
Taking a left limit in the first inequality also gives $F(y)-v_+\le z\sqrt{f(y)}$. 

Decompose $x\in[0,B]$ as the sum of three nondecreasing, $1$-Lipschitz clipped parts,
\[
 L(x)=x\wedge\ell,\quad
 J(x)=((x-\ell)\vee0)\wedge(B-2\ell),\quad
 R(x)=(x-B+\ell)\vee0,
 \qquad x=L(x)+J(x)+R(x).
\]
If $q(v_+)<y\wedge\ell$, let $b=F(x)-v_+>0$ for a point between these two locations. Then $x=q(v_++b)$ and \eqref{eq:shifted-cdf-relations} gives
\[
 g(v_++b)\le z^2/b^2.
\]
Boundary monotonicity also gives $g(v_++b)\le g(v_+)$. Thus, writing $y_L=y\wedge\ell$,
\begin{align*}
 L(y)-L(q(v_+))
 &=\int_0^{F(y_L)-v_+}g(v_++b)\,\rd b\\
 &\le\int_0^\infty\min\{g(v_+),z^2/b^2\}\,\rd b
 =2z\sqrt{g(v_+)}.
\end{align*}
Consequently,
\[
 [L(y)-L(q(v_+))]_+^2\le H_{L,+}(v_+).
\]

For the other direction, suppose $L(y)<L(q(v_-))$. Then $y<\ell$ and \eqref{eq:shifted-cdf-relations} implies $A_z(y)\ge v_-$. In particular $v_-\le A_z(\ell)$ and $y\ge a=a_z(v_-)$. Because $A_z(a)=v_-$, monotonicity of $f$ on $[a,\ell]$ gives
\[
 f(a)\{L(q(v_-))-a\}
 \le F(L(q(v_-)))-F(a)
 \le v_--F(a)=z\sqrt{f(a)}.
\]
Since $0\le L\le\ell$, the preceding bound gives
\[
 [L(q(v_-))-L(y)]_+^2\le\min\left\{\ell^2,\frac{z^2}{f(a_z(v_-))}\right\}=H_{L,-}(v_-).
\]
Using $q(v_-)\le q(u)\le q(v_+)$ and the
$1$-Lipschitz property of $L$, we obtain
\begin{align*}
 L(y)-L(q(u))&\le r(u)+\sqrt{H_{L,+}(v_+)},\\
 L(q(u))-L(y)&\le r(u)+\sqrt{H_{L,-}(v_-)}.
\end{align*}
Consequently,
\[
 |L(y)-L(q(u))|\le r(u)+\sqrt{H_{L,+}(v_+)}+\sqrt{H_{L,-}(v_-)}.
\]

The reflected construction gives the same bound for $R$. Since $v_-\le v_+$ and the left and right functions are respectively nonincreasing and nondecreasing,
\[
 \begin{aligned}
 H_{L,+}(v_+)+H_{L,-}(v_-)&\le H_L(v_-),\\
 H_{R,+}(v_+)+H_{R,-}(v_-)&\le H_R(v_+).
 \end{aligned}
\]

Since $f\ge c_{\gM,\ell}$ on $[\ell,B-\ell]$, \eqref{eq:shifted-cdf-relations} and its left limit give
\begin{align*}
 c_{\gM,\ell}[J(y)-J(q(v_+))]_+&\le [F(y)-v_+]_+\le z\sqrt{f(y)}\le z\sqrt{C_0},\\
 c_{\gM,\ell}[J(q(v_-))-J(y)]_+&\le [v_--F(y)]_+\le z\sqrt{f(y)}\le z\sqrt{C_0}.
\end{align*}
As $q(v_-)\le q(u)\le q(v_+)$ and $J$ is $1$-Lipschitz,
\[
 |J(y)-J(q(u))|\le r(u)+\frac{z\sqrt{C_0}}{c_{\gM,\ell}}.
\]
Adding the three clipped bounds proves \eqref{eq:shifted-pointwise} with a constant depending only on $\gM$ and $\ell$. If $z=0$, both functions $H_L,H_R$ may then be set to zero, proving that case as well.
\end{proof}

\begin{lemma}\label{lem:shifted-grid}
Under Assumption~\ref{ass:A4}, let $\ell$ and $\rho_0=\rho_0(\gM,\ell)$ be as in Lemma~\ref{lem:common-localization}. Fix $m\ge1$, $0<\rho\le\rho_0$, $s\in\gS$ and $\btheta\in\gU_m(\rho)$, and write $f=f_{s,\btheta}$ and $q=F_{s,\btheta}^{-1}$. If $0\le\varepsilon<1/(4m)$, then for every nonnegative monotone function $H$ on $(0,1)$,
\[
 \frac1m\sum_{i=1}^mH(\tau_i+\varepsilon)
 \le4\int_0^1H(v)\,\rd v,\quad \frac1m\sum_{i=1}^mH(\tau_i-\varepsilon)
 \le4\int_0^1H(v)\,\rd v
\]
With $r(u)=q(u+\varepsilon)-q(u-\varepsilon)$,
\begin{equation}\label{eq:rank-shift-average}
 \frac1m\sum_i r(\tau_i)^2
 \le B\prn{16\ell+\frac{2}{c_{\gM,\ell}}}\varepsilon.
\end{equation}
\end{lemma}
\begin{proof}
The shifted levels $w_i=\tau_i\pm\varepsilon$ are at distance at least $1/(4m)$ from both endpoints of their original cells. If $H$ is nonincreasing, then
\[
 \frac1m H(w_i)\le4\int_{(i-1)/m}^{w_i}H(v)\,\rd v.
\]
Summing over the disjoint cells gives the first assertion. For nondecreasing $H$, integrate from $w_i$ to $i/m$ instead.

By Lemma~\ref{lem:common-localization} and \eqref{eq:boundary-mono}, $g(v)=q'(v)=1/f(q(v))$, and
\[
 r(\tau_i)=\int_{\tau_i-\varepsilon}^{\tau_i+\varepsilon}g(v)\,\rd v.
\]
Since $\varepsilon<1/(4m)$, we have
\[
 \tau_i-\varepsilon-\frac{i-1}{m}\ge\frac1{4m},
 \qquad
 \frac{i}{m}-\tau_i-\varepsilon\ge\frac1{4m}.
\]
By boundary monotonicity, for
$v\in[\tau_i-\varepsilon,\tau_i+\varepsilon]$,
\begin{align*}
 g(v)\ind\{q(v)<\ell\}
 &\le4m\int_{(i-1)/m}^{\tau_i-\varepsilon}
 g(w)\ind\{q(w)<\ell\}\,\rd w,\\
 g(v)\ind\{q(v)>B-\ell\}
 &\le4m\int_{\tau_i+\varepsilon}^{i/m}
 g(w)\ind\{q(w)>B-\ell\}\,\rd w.
\end{align*}
Consequently,
\begin{align*}
 &\frac1m\sum_i\int_{\tau_i-\varepsilon}^{\tau_i+\varepsilon}
 g(v)\ind\{q(v)<\ell\}\,\rd v\le8\varepsilon\int_{\{q(w)<\ell\}}g(w)\,\rd w\le8\ell\varepsilon,\\
 &\frac1m\sum_i\int_{\tau_i-\varepsilon}^{\tau_i+\varepsilon}g(v)\ind\{\ell\le q(v)\le B-\ell\}\,\rd v\le\frac{2\varepsilon}{c_{\gM,\ell}},\\
 &\frac1m\sum_i\int_{\tau_i-\varepsilon}^{\tau_i+\varepsilon}g(v)\ind\{q(v)>B-\ell\}\,\rd v\le8\varepsilon\int_{\{q(w)>B-\ell\}}g(w)\,\rd w\le8\ell\varepsilon.
\end{align*}
Summing these bounds yields
\[
 \frac1m\sum_i r(\tau_i)\le(16\ell+2/c_{\gM,\ell})\varepsilon.
\]
Equation~\eqref{eq:rank-shift-average} follows from the fact that $0\le r(v)\le B$.
\end{proof}

To prove Lemma~\ref{lem:joint-bernstein-event}, we first bound the expected error of an empirical reward CDF after weighting by the density, and then obtain a concentration bound by truncating the weight. 

\begin{lemma}\label{lem:weighted-doob}
Let $(M_u)_{0\le u\le T}$ be a square-integrable c\`adl\`ag martingale with $M_0=0$, and let $w:(0,T]\to[0,\infty)$ be continuous and nonincreasing. Then, allowing either side to be infinite,
\[
 \E\sup_{0<u\le T}w(u)M_u^2
 \le4\int_{(0,T]}w(u)\,\rd\E M_u^2.
\]
\end{lemma}
\begin{proof}
First suppose $w$ is bounded, extended at zero by its right limit, and put $M_u^*=\sup_{0\le v\le u}|M_v|$. For $u\le T$,
\[
 w(u)=w(T)+\int_{(u,T]}(-\rd w(v)).
\]
Since $M_u^2\le(M_v^*)^2$ for $v\ge u$ and $-\rd w$ is a positive measure,
\[
 \sup_{u\le T}w(u)M_u^2
 \le w(T)(M_T^*)^2+\int_{(0,T]}(M_v^*)^2(-\rd w(v)).
\]
By Doob's maximal inequality,
\[
 \E\sup_{u\le T}w(u)M_u^2\leq4\left[w(T)\E M_T^2+\int_{(0,T]}\E M_v^2(-\rd w(v))\right].
\]
Integration by parts gives
\[
 w(T)\E M_T^2+\int_{(0,T]}\E M_v^2(-\rd w(v))=\int_{(0,T]}w(v)\,\rd \E M_v^2,
\] 
which proves the claim for bounded $w$. For general $w$, apply the argument above to $w_K=w\wedge K$ and let $K\to\infty$.
\end{proof}

\begin{lemma}\label{lem:weighted-ecdf}
Let $F$ be a distribution on $[0,1]$ with bounded density $f$.  Suppose $f$ is continuous and positive on $(0,1)$, nondecreasing on $[0,\kappa]$, nonincreasing on $[1-\kappa,1]$. For $N$ i.i.d. samples with empirical CDF $F_N$, define
\[
  Z_N(F)\coloneq \sup_{x:\,f(x)>0}
  \frac{\abs{F_N(x)-F(x)}}{\sqrt{f(x)}}.
\]
Then
\begin{equation}\label{eq:weighted-ecdf-L2}
  \E Z_N(F)^2\le\frac{C(F)}N.
\end{equation}
\end{lemma}
\begin{proof}
Let $U_1,\ldots,U_N$ be iid variables with uniform distribution on $[0,1]$. Denote the empirical CDF by $U_N$, and write $q=F^{-1}$.
By the probability integral transform,
\[
 Z_N(F)=\sup_{0<u<1}\frac{|U_N(u)-u|}{\sqrt{f(q(u))}}.
\]
Choose $0<\kappa_0<\min\{\kappa,1/2\}$ and set $T=F(\kappa_0)<1$. For $0\le u\le T$, define
\[
 \gF_u=\sigma\prn{\ind\{U_k\le v\}: k\in[N],\ 0\le v\le u}.
\] 
Then the process
\[
 \prn{M_u\coloneq\frac{U_N(u)-u}{1-u}}_{0\le u\leq T}
\]
is a square-integrable martingale with respect to $(\gF_u)_{0\le u\le T}$. Direct calculation gives
\[
 \E M_u^2=\frac{u}{N(1-u)},\qquad
 \rd\E M_u^2=\frac{\rd u}{N(1-u)^2}.
\]
On the left boundary, the weight $w(u)=(1-u)^2/f(q(u))$ is nonincreasing because $f(q(u))$ is nondecreasing. Lemma~\ref{lem:weighted-doob} therefore gives
\begin{align*}
 \E\sup_{0<u\le T}\frac{|U_N(u)-u|^2}{f(q(u))}
 &\le4\int_0^T\frac{(1-u)^2}{f(q(u))}\frac{\rd u}{N(1-u)^2}\\
 &=\frac4N\int_0^T\frac{\rd u}{f(q(u))}
 =\frac{4\kappa_0}{N}.
\end{align*}
The last equality is the change of variables $u=F(x)$ on the active reward support. Reflection gives the same bound for the right boundary. Finally, let $c_F\coloneq \min_{x\in[\kappa_0,1-\kappa_0]} f(x)$. DKW inequality gives
\[
 \E\sup_{x\in[\kappa_0,1-\kappa_0]}\frac{|F_N(x)-F(x)|^2}{f(x)}\le \frac1{c_FN}\E\sup_x|F_N(x)-F(x)|^2\le\frac{2}{c_FN}.
\]
Adding the three bounds proves \eqref{eq:weighted-ecdf-L2}. 
\end{proof}

\begin{lemma}\label{lem:weighted-ecdf-hp}
Under the assumptions of Lemma~\ref{lem:weighted-ecdf}, there is a finite constant $C_F$ such that, for every integer $N\ge1$ and every $t\ge1$, with probability at least $1-e^{-t}$, for every $x\in[0,1]$,
\begin{equation}\label{eq:weighted-ecdf-hp}
 |F_N(x)-F(x)|\le C_F\left\{\sqrt{\frac{t f(x)}{N}}+\frac{t}{N}\right\},
\end{equation}
where $C_F$ depends only on $F$.
\end{lemma}
\begin{proof}
Let $\tau=t/N$ and put
\[
 S_\tau=\sup_{x\in[0,1]}
 \frac{|F_N(x)-F(x)|}{\sqrt{f(x)\vee\tau}}.
\]
By Lemma~\ref{lem:weighted-ecdf} and Jensen's inequality,
\begin{equation}\label{eq:truncated-mean}
 \E S_\tau\le \bigl(\E Z_N(F)^2\bigr)^{1/2}\le C_F N^{-1/2}.
\end{equation}
Boundary monotonicity gives $F(x)/f(x)\le x$ near zero and $(1-F(x))/f(x)\le1-x$ near one. Consequently,
\begin{equation*}
 \sigma_F^2\coloneq \sup_{0<x<1}\frac{F(x)(1-F(x))}{f(x)}<\infty.
\end{equation*}
Consider the function class
\[
 \gH_\tau=\left\{r\mapsto\frac{\pm(\ind\{r\le x\}-F(x))}{\sqrt{f(x)\vee\tau}}:x\in[0,1]\right\}.
\]
Then for iid samples $R_1,\ldots,R_N$ with distribution $F$, we have
\[
 S_\tau=\sup_{h\in\gH_\tau}\frac{1}{N}\sum_{n=1}^Nh(R_n).
\]
For every $h\in\gH_\tau$, $|h|\leq\tau^{-1/2}$, $\E h(R)=0$ and $\var(h(R))\leq \sigma_F^2$. Apply Bousquet's inequality \citep{boucheron2013concentration} and we have
\[
 \PB\left\{S_\tau>\E S_\tau+
 \sqrt{\frac{2(\sigma_F^2+2\tau^{-1/2}\E S_\tau)t}{N}}+\frac{\tau^{-1/2}t}{3N}\right\}
 \le e^{-t}.
\]
Therefore,
\begin{equation}\label{eq:bousquet-truncated}
 \PB\left\{S_\tau>2\E S_\tau+
 C(\sigma_F+1)\sqrt{\frac{t}{N}}\right\}
 \le e^{-t},
\end{equation}
where $C$ is universal. Since $t\ge1$, \eqref{eq:truncated-mean}--\eqref{eq:bousquet-truncated} show that 
\begin{equation*}
    \PB\brc{S_\tau\leq C_F\sqrt{\frac tN}}\geq 1-e^{-t}.
\end{equation*}
On this event, for every $x$,
\[
 |F_N(x)-F(x)|\le C_F\sqrt{\frac tN}\sqrt{f(x)\vee\frac tN}
 \le C_F\left(\sqrt{\frac{tf(x)}N}+\frac tN\right).
\]
This proves \eqref{eq:weighted-ecdf-hp}.
\end{proof}

%% file: appendices/local.tex
For $Q\in\gQ$, define
\[
 J_{Q,s,a,s^\prime}(x)=Q_{s,a}(({-\infty},x]\times\{s^\prime\}),
 \qquad
 D_n(Q)=\max_s\sum_a\pi(a\mid s)\sum_{s^\prime}
 \norm{\what J_{s,a,s^\prime}-J_{Q,s,a,s^\prime}}_\infty.
\]

\begin{lemma}\label{lem:local-moments}
There exists a constant $L_{m,Q_0}<\infty$ depending only on $m$ and $Q_0$, such that
\begin{equation}\label{eq:local-pop-stability}
 \norm{\btheta_m(Q)-\btheta_m(Q_0)}_\infty\le L_{m,Q_0}d_H(Q,Q_0).
\end{equation}
Moreover, for fixed $c<\infty$ and $r\ge1$,
\begin{equation}\label{eq:local-uniform-moments}
 \sup_{n\ge1}\sup_{Q\in\gN_n(c;Q_0)}
 \E_Q\left[\left\{\sqrt n\,
 \norm{\btheta_m^{(n)}-\btheta_m(Q)}_\infty\right\}^r\right]
 \le C_{m,c,r,Q_0}<\infty.
\end{equation}
\end{lemma}
\begin{proof}
Write $\btheta_0=\btheta_m(Q_0)$ and
$F_{0,s}=F_{[\gT_{Q_0}^\pi\Beta_{\btheta_0}](s)}$, with density $f_{0,s}$.
By Lemma~A.2 in~\citet{cheng2026statistical}, there exist $\kappa,h>0$ such that,
for all $(s,i)$ and $|x-\theta_0(s,i)|\le h$, $f_{0,s}(x)\ge\kappa$.
Therefore, for $q=\theta_0(s,i)$ and $0\le v\le h$,
\[
 F_{0,s}(q-v)\le\tau_i-\kappa v,\qquad F_{0,s}(q+v)\ge\tau_i+\kappa v.
\]
For any CDF $G_s$ supported on $I$, put
$e=\|G_s-F_{0,s}\|_\infty$ and we have
\[
 \abs{G_s^{-1}(\tau_i)-q}\le\begin{cases}
 e/\kappa,&e<\kappa h/2,\\
 B,&e\ge\kappa h/2,
 \end{cases}
 \le C_{m,Q_0}e,
 \qquad C_{m,Q_0}=\max\left\{\frac1\kappa,\frac{2B}{\kappa h}\right\}.
\]

By the definition of total variation and \eqref{eq:joint-cdf-decomp}, applied under $Q$, we have, uniformly in $\btheta$,
\begin{align*}
 \max_s\norm{F_{[\gT_Q^\pi\Beta_{\btheta}](s)}-F_{[\gT_{Q_0}^\pi\Beta_{\btheta}](s)}}_\infty&\le\max_b\TV(Q_b,Q_{0,b})\le d_H(Q,Q_0),\\
 \max_s\norm{\what F_{s,\btheta}^{(n)}-F_{[\gT_Q^\pi\Beta_{\btheta}](s)}}_\infty&\le D_n(Q).
\end{align*}
For $Q\in\gQ$, define
\[
 [\Phi_{m,Q}(\btheta)]_{s,i}=F_{[\gT_Q^\pi\Beta_{\btheta}](s)}^{-1}(\tau_i).
\]
Proposition~\ref{prop:basic_properties} implies
\begin{align*}
 (1-\gamma)\norm{\btheta_m(Q)-\btheta_0}_\infty&\le\norm{\Phi_{m,Q}(\btheta_0)-\btheta_0}_\infty\le C_{m,Q_0}d_H(Q,Q_0),\\
 (1-\gamma)\norm{\btheta_m^{(n)}-\btheta_0}_\infty&\le\norm{\what\Phi_{m,n}(\btheta_0)-\btheta_0}_\infty\le C_{m,Q_0}\{D_n(Q)+d_H(Q,Q_0)\}.
\end{align*}
Therefore \eqref{eq:local-pop-stability} holds with $L_{m,Q_0}=C_{m,Q_0}/(1-\gamma)$. In particular,
\[
 \norm{\btheta_m^{(n)}-\btheta_m(Q)}_\infty
 \le L_{m,Q_0}\{D_n(Q)+2d_H(Q,Q_0)\}.
\]

The DKW argument in the proof of Lemma~\ref{lem:primitive-conc},
applied under $Q$ and followed by a union bound over all $|\gS|^2|\gA|$ triples, gives
\[
 \PB_Q\{D_n(Q)>t\}
 \le 2|\gS|^2|\gA|\exp\left(-\frac{2nt^2}{|\gS|^2}\right).
\]
Integrating this tail bound yields, for every $r\ge1$,
\[
 \sup_{Q\in\gQ}\E_Q[(\sqrt nD_n(Q))^r]
 \le C_r|\gS|^r\{1+\log(2|\gS|^2|\gA|)\}^{r/2}.
\]
Since $\sqrt n\,d_H(Q,Q_0)\le c$ on $\gN_n(c;Q_0)$,
\eqref{eq:local-uniform-moments} follows.
\end{proof}

\begin{lemma}\label{lem:local-increments}
Let $\btheta_0=\btheta_m(Q_0)$. There exists $C_{m,Q_0}<\infty$, depending only on $m$ and $Q_0$, such that for every $\delta\in(0,1)$ and $Q\in\gQ$,
\begin{align}
 &\E_Q\sup_{\|\btheta-\btheta_0\|_\infty\le\delta}\sqrt n\left\|(\bH_n-\bH_Q)(\btheta)-(\bH_n-\bH_Q)(\btheta_0)\right\|_\infty\le C_{m,Q_0}\sqrt{\delta+d_H(Q,Q_0)^2},\label{eq:local-emp-increment}\\
 &\sup_{\|\btheta-\btheta_0\|_\infty\le\delta}\left\|(\bH_Q-\bH_{Q_0})(\btheta)-(\bH_Q-\bH_{Q_0})(\btheta_0)\right\|_\infty\le C_{m,Q_0}d_H(Q,Q_0)\sqrt{\delta+d_H(Q,Q_0)^2}.\label{eq:local-pop-increment}
\end{align}
\end{lemma}

\begin{proof}
Fix $s,s^\prime\in\gS$, $a\in\gA$ and $i,j\in[m]$, and write $b=(s,a)$. Set
\[
 x_0=\theta_0(s,i)-\gamma\theta_0(s^\prime,j),
 \qquad
 A_\delta=\{(r,s^\prime):|r-x_0|\le(1+\gamma)\delta\}.
\]
For $\|\btheta-\btheta_0\|_\infty\le\delta$, the corresponding threshold $x=\theta(s,i)-\gamma\theta(s^\prime,j)$ satisfies $|x-x_0|\le(1+\gamma)\delta$. By the Hellinger inequality,
\begin{align*}
 Q_{0,b}(A_\delta)&\le2(1+\gamma)C_0\delta,\\
 Q_b(A_\delta)&\le\left[\sqrt{Q_{0,b}(A_\delta)}+d_H(Q,Q_0)\right]^2\\
 &\le C_{m,Q_0}\brk{\delta+d_H(Q,Q_0)^2}.
\end{align*}

For $v\in\RB$, let $B_v=A_\delta\cap\{(r,s^\prime):r\le v\}$, and put $p_A=Q_b(A_\delta)$ and
\[
 N_A=\sum_{k=1}^n\ind\{X_{k,b}\in A_\delta\}.
\]
If $p_A=0$, the bound is immediate, hence we assume $p_A>0$. Let $F_A$ be the conditional CDF of the reward given $A_\delta$, and $F_{N_A,A}$ its empirical CDF. Then
\[
(\what Q_b^{(n)}-Q_b)(B_v)=\frac{N_A}{n}\{F_{N_A,A}(v)-F_A(v)\}+\left(\frac{N_A}{n}-p_A\right)F_A(v),
\]
where the first term is zero if $N_A=0$.
Conditional DKW inequality and $N_A\sim\operatorname{Bin}(n,p_A)$ give
\begin{align*}
&\sqrt n\,\E_Q\sup_v|(\what Q_b^{(n)}-Q_b)(B_v)|\\
\leq&\frac C{\sqrt n}\E_Q\sqrt{N_A}+\sqrt n\,\E_Q\left|\frac{N_A}{n}-p_A\right|\\
\leq& C\sqrt{p_A}\le C_{m,Q_0}\sqrt{\delta+d_H(Q,Q_0)^2}.
\end{align*}

For $|x-x_0|\le(1+\gamma)\delta$,
\[
 \ind\{t=s^\prime,r\le x\}
 -\ind\{t=s^\prime,r\le x_0\}
 =\ind_{B_x}(r,t)-\ind_{B_{x_0}}(r,t).
\]
Consequently,
\[
\sup_{|x-x_0|\le(1+\gamma)\delta}\abs{(\what Q_b^{(n)}-Q_b)(B_x)-(\what Q_b^{(n)}-Q_b)(B_{x_0})}\le2\sup_v\abs{(\what Q_b^{(n)}-Q_b)(B_v)}.
\]
Summing over the finitely many indices proves
\eqref{eq:local-emp-increment}.

For the population increment, Cauchy inequality yields
\begin{align*}
&\sup_{|x-x_0|\le(1+\gamma)\delta}\abs{(Q_b-Q_{0,b})(B_x)-(Q_b-Q_{0,b})(B_{x_0})}\\
\leq&d_H(Q,Q_0)\sqrt{2\brc{Q_b(A_\delta)+Q_{0,b}(A_\delta)}}\\
\leq&C_{m,Q_0}d_H(Q,Q_0)\sqrt{\delta+d_H(Q,Q_0)^2}.
\end{align*}
Summing over the same indices proves \eqref{eq:local-pop-increment}.
\end{proof}

%% file: appendices/construction.tex
This appendix supplies the technical construction used in the proofs of Lemmas~\ref{lem:hard-rewards} and~\ref{lem:hard-resolutions} in Appendix~\ref{app:proof-hard-instance-group}. The MRP has two states $\{s,t\}$, one action, $P(t\mid s)=P(t\mid t)=1$, and $\gamma=1/2$. Throughout this appendix, $c,C,C',\ldots$ denote positive constants depending on the fixed construction. They may change from one line to the next and are independent of $j,m,n,p$, unless some additional dependence is stated.

\begin{lemma}\label{lem:generic}
Let $p_s^0$ be the shelf density constructed in the proof of Lemma~\ref{lem:hard-rewards}, and let $p_t^0(r)=6r(1-r)$. There are densities $p_s,p_t$ that satisfy Assumptions~\ref{ass:A1},~\ref{ass:A2}, and~\ref{ass:A4} for all resolutions, with
\[
 \frac34 p_u^0(r)\le p_u(r)\le\frac54 p_u^0(r),
 \qquad u\in\{s,t\},\quad r\in[0,1].
\]
\end{lemma}
\begin{proof}
Choose $\kappa>0$ such that $(p_u^0)'$ is strictly positive on $(0,\kappa]$ and strictly negative on $[1-\kappa,1)$, and put $E=(0,\kappa)\cup(1-\kappa,1)$. Define
\begin{equation*}
    \gH=\brc{h=(h_s,h_t)\in (C^1[0,1])^2:h_u(0)=h_u(1)=0, \int_0^1 h_u=0\text{ for }u=s,t; \|h\|_*<\infty},
\end{equation*}
where
\[
 \|h\|_*=\max_{u\in\{s,t\}}\left(\|h_u\|_{C^1}+\sup_{0<r<1}\frac{|h_u(r)|}{p_u^0(r)}+\sup_{r\in E}\frac{|h_u'(r)|}{|(p_u^0)'(r)|}\right).
\]
It is easy to verify that the space $(\mathcal H,\|\cdot\|_*)$ is Banach.
On the open ball $\gB=\{h\in\gH:\|h\|_*<1/4\}$, the densities $p_u=p_u^0+h_u$ obey
\[
 \frac34p_u^0\le p_u\le\frac54p_u^0,
 \qquad \operatorname{sgn}(p_u')=\operatorname{sgn}((p_u^0)')\quad\text{on }E.
\]
They are therefore positive on $(0,1)$ and satisfy Assumptions~\ref{ass:A1} and~\ref{ass:A4}.

Fix $m$ and $h\in\gB$. We show that each equality $\theta_m(u,i)-\gamma\theta_m(u',j)\in\{0,1\}$ can be broken by an arbitrarily small perturbation. Set $z_{u,i,j}=\theta_m(u,i)-\gamma\theta_m(t,j)$. For $u=s,t$, choose $H_u\in C_c^\infty(0,1)$ such that
\[
 H_u(z_{u,i,j})=-p_u(z_{u,i,j})\quad\text{if }z_{u,i,j}\in(0,1).
\]
Such functions exist by interpolation at finitely many points. The equality also holds outside $(0,1)$, where both functions vanish.

For sufficiently small $|\epsilon|$, $h+\epsilon(H_s',H_t')\in\gB$. The corresponding densities $p_u^\epsilon=p_u+\epsilon H_u'$ have CDFs $F_u^\epsilon=F_u+\epsilon H_u$.
Writing $z_{u,i,j}^\epsilon=\theta_m^\epsilon(u,i)-\gamma\theta_m^\epsilon(t,j)$, the perturbed fixed-point equations are
\[
 \frac1m\sum_{j=1}^m\left[F_u(z_{u,i,j}^\epsilon)+\epsilon H_u(z_{u,i,j}^\epsilon)\right]=\tau_i.
\]
For $u=s,t$, we have
\[
 f_{m,u}(\theta_m(u,i))=\frac1m\sum_{j=1}^m p_u(z_{u,i,j})>0,
\]
since otherwise the fixed-point equation would imply $\tau_i\in\{0,1/m,\ldots,1\}$. According to the notations at the beginning of Appendix~\ref{app:proof-local-minimax-group}, $\bG_m=\bD_m(\bI-\gamma\bK_m)$ is invertible, and $\bG_m\mathbf1=(1-\gamma)\bD_m\mathbf1$. The implicit function theorem therefore gives the differentiability of $\btheta_m^\epsilon$ at $\epsilon=0$.
We differentiate the perturbed fixed-point equations to obtain
\[
\begin{aligned}
[\bG_m\dot\btheta_m]_{u,i}
&=-\frac1m\sum_{j=1}^m H_u(z_{u,i,j})=\frac1m\sum_{j=1}^m p_u(z_{u,i,j})\\
&=[\bD_m\mathbf1]_{u,i}=[(1-\gamma)^{-1}\bG_m\mathbf1]_{u,i}.
\end{aligned}
\]
Consequently, $\dot\btheta_m=(1-\gamma)^{-1}\mathbf1$ and for all $u,u'\in\{s,t\}$ and $i,j$,
\[
 \left.\frac{\rd}{\rd\epsilon}\left[\theta_m^\epsilon(u,i)-\gamma\theta_m^\epsilon(u',j)\right]\right|_{\epsilon=0}=1.
\]

For each fixed $m,u,u',i,j$ and $c\in\{0,1\}$, the set $\{h\in\gB:\theta_m^h(u,i)-\gamma\theta_m^h(u',j)=c\}$ is therefore closed and has empty interior relative to the ball, where $\btheta_m^h$ is the fixed point corresponding to $p^0+h$. By the Baire category theorem, the set
\begin{equation*}
    \gB^0\coloneq\gB\setminus\bigcup_{\substack{m,u,u^\prime,i,j,\\ c\in\{0,1\}}}\{h\in\gB:\theta_m^h(u,i)-\gamma\theta_m^h(u',j)=c\}\neq\varnothing.
\end{equation*}
Choosing $h^0\in\gB^0$ gives a density pair $(p_s^0+h_s^0,p_t^0+h_t^0)$ satisfying Assumption~\ref{ass:A2} and the required bounds.
\end{proof}

\begin{lemma}\label{lem:hard-localization}
With the notation of Section~\ref{subsec:proof-instances},
for all sufficiently large $j$,
\[
 c\epsilon_jx_j e^{-Cj^2}\le a_j\le C\epsilon_jx_j,\qquad
 q_j\in[x_j/2,x_j],\qquad
 \nu([0,x_j/8])\ge e^{-Cj^2}.
\]
\end{lemma}
\begin{proof}
We first show that
\[
 \nu([0,x_j/8])\ge e^{-Cj^2}.
\]
Let $R_{t,k}$ be the independent rewards at state $t$, and set $K_j=\lceil\log_2(32/x_j)\rceil=O(j)$. On the event $\{R_{t,k}\le x_j/32,\ 0\le k<K_j\}$,
\[
 \sum_{k=0}^\infty 2^{-k}R_{t,k} \le \frac{x_j}{16}+\sum_{k=K_j}^\infty2^{-k}\le\frac{x_j}{8}.
\]
Since $p_t(r)\asymp r$ near zero,
\[
 \nu([0,x_j/8])\ge\bigl[\PB(R_t\le x_j/32)\bigr]^{K_j}\ge(c x_j^2)^{K_j}\ge e^{-Cj^2}.
\]
Therefore, by Lemma~\ref{lem:hard-rewards},
\[
\begin{aligned}
 a_j&=\E F_{R_s}(3x_j/4-\gamma Z)\ge c\epsilon_jx_j\,\nu([0,x_j/8])\ge c\epsilon_jx_j e^{-Cj^2},\\
 a_j&\le F_{R_s}(3x_j/4)\le C\epsilon_jx_j.
\end{aligned}
\]

By \eqref{eq:basic-approx}, $W_1(\nu_m,\nu)\le C/m$. Since $p_s(r)\le C\epsilon_j$ for $r\le x_j$,
\[
 \sup_{x\le x_j}|H_m(x)-H(x)|\le C\epsilon_j W_1(\nu_m,\nu)\le\frac{C\epsilon_j}{m}.
\]
Moreover, for $x\in[x_j/2,x_j]$,
\[
 H'(x)=\E p_s(x-\gamma Z)\ge c\epsilon_j\,\nu([0,x_j/8])\ge c\epsilon_j e^{-Cj^2}.
\]
Consequently,
\[
 \min\left\{H(3x_j/4)-H(x_j/2),H(x_j)-H(3x_j/4)\right\}\ge c\epsilon_jx_j e^{-Cj^2}.
\]
Since $m_j\asymp a_j^{-1}$ and
\[
 \left|\frac1{2m_j}-a_j\right|\le Ca_j^2,
 \qquad
 \frac{a_j^2+\epsilon_j/m_j}{\epsilon_jx_j e^{-Cj^2}}\le C\epsilon_j e^{Cj^2}\to 0,
\]
we obtain that for all sufficiently large $j$,
\[
 H_{m_j}(x_j/2)<\frac1{2m_j}<H_{m_j}(x_j).
\]
Hence $q_j\in[x_j/2,x_j]$.
\end{proof}